\documentclass{article}

\usepackage[final, nonatbib]{neurips_2022}

\usepackage[utf8]{inputenc} 
\usepackage[T1]{fontenc}    
\usepackage{hyperref}       
\usepackage{url}            
\usepackage{booktabs}       
\usepackage{amsfonts}       
\usepackage{nicefrac}       
\usepackage{microtype}      
\usepackage{verbatim}

\usepackage{tabularx}
\usepackage{tabulary}
\usepackage{booktabs}

\usepackage[sort, numbers]{natbib}
\usepackage{xcolor}
\definecolor{navyblue}{rgb}{0.0, 0.0, 0.5}
\hypersetup{colorlinks, citecolor=blue}

\usepackage{graphicx}
\usepackage{subcaption}
\usepackage{mwe}
\usepackage{wrapfig}
\usepackage{pifont}

\usepackage{enumitem}

\usepackage{algorithmic}
\usepackage{algorithm}
\usepackage{tikz}
\usetikzlibrary{fit,calc}

\usepackage{amssymb}
\usepackage{amsmath}
\usepackage{amsthm}
\usepackage{bbm}
\usepackage{wrapfig,lipsum,booktabs}
\usepackage{dsfont}
\usepackage{mathtools}
\usepackage{mathabx}
\newlength{\depthofsumsign}
\newcommand{\nsum}[1][1.4]{
    \mathop{%
        \raisebox
            {-#1\depthofsumsign+1\depthofsumsign}
            {\scalebox
                {#1}
                {$\displaystyle\sum$}%
            }
    }
}

\newlength{\depthofmathbbEsign}
\newtheorem{definition}{Definition}

\usepackage{commath}
\usepackage{relsize}

\usepackage{multirow}
\usepackage{multicol}
\usepackage{subcaption}
\usepackage{caption, makecell}
\usepackage{array}
\usepackage{arydshln}
\usepackage{booktabs}
\usepackage{colortbl}
\usepackage{color}
\usepackage{tablefootnote}
\usepackage{hhline}
\usepackage{threeparttable}

\usepackage{CJKutf8}

\usepackage{comment}

\makeatletter
\def\thanks#1{\protected@xdef\@thanks{\@thanks
        \protect\footnotetext{#1}}}
\makeatother

\usepackage{ifpdf}

\theoremstyle{plain}

\newtheorem{proposition}{Proposition}

\theoremstyle{definition}

\title{Preservation of the Global Knowledge by \\ Not-True Distillation in Federated Learning}

\author{%
  Gihun Lee*, Minchan Jeong*, Yongjin Shin, Sangmin Bae, Se-Young Yun\\
  KAIST\\
    \texttt{\{opcrisis, mcjeong, yj.shin, bsmn0223, yunseyoung\}@kaist.ac.kr}\thanks{* Equal contribution.}
}

\begin{document}
\begin{CJK}{UTF8}{mj} 
\maketitle

\begin{abstract}
In federated learning, a strong global model is collaboratively learned by aggregating clients' locally trained models. Although this precludes the need to access clients' data directly, the global model's convergence often suffers from data heterogeneity. This study starts from an analogy to continual learning and suggests that \textit{forgetting} could be the bottleneck of federated learning. We observe that the global model forgets the knowledge from previous rounds, and the local training induces forgetting the knowledge outside of the local distribution. Based on our findings, we hypothesize that tackling down forgetting will relieve the data heterogeneity problem. To this end, we propose a novel and effective algorithm, \textit{Federated Not-True Distillation} (FedNTD), which preserves the global perspective on locally available data only for the \textit{not-true} classes. In the experiments, FedNTD shows state-of-the-art performance on various setups without compromising data privacy or incurring additional communication costs\footnote{\href{https://github.com/Lee-Gihun/FedNTD}{https://github.com/Lee-Gihun/FedNTD}}.
\end{abstract}

\section{Introduction}
\vspace{-6pt}

At present, massive data is being collected from edge devices such as mobile phones, vehicles, and facilities. As the data may be distributed on numerous devices, decentralized training is often required to train deep network models. Federated learning \citep{federated_learning, federated_optimization} is a distributed learning paradigm that enables the learning of a global model while preserving clients' data privacy. In federated learning, clients independently train local models using their private data, and the server aggregates them into a single global model. In this process, most of the computation is performed by client devices, while the global server only aggregates the model parameters and distributes them to clients \citep{fl_survey_technologies_applications, federatedML_concepts_applications}.

Most federated learning algorithms are based on FedAvg \citep{FedAvg}, which aggregates the locally trained model parameters by weighted averaging proportional to the amount of local data that each client had. While various federated learning algorithms have been proposed thus far, they each conduct parameter averaging in a certain manner \citep{FedProx, SCAFFOLD, FedNova, FedBE, FedMix, FedDyn}. Although this aggregation scheme empirically works well and provides a conceptually ideal framework when all client devices are active and i.i.d. distributed (a.k.a. LocalSGD), the data heterogeneity problem \citep{On_the_convergence_fedavg_noniid, fl_with_noniid} is a notorious challenge for federated learning applications and prevents their widespread applicability \citep{fl_challenges_methods, advances_open_problems_fl}.

As the clients generate their own data, the data is not identically distributed. More precisely, the local data across clients are drawn from heterogeneous underlying distributions; thereby, locally available data fail to represent the overall global distribution, which is referred to as data heterogeneity. Despite its inevitable occurrence in many real-world scenarios, data heterogeneity not only makes theoretical analysis difficult \citep{On_the_convergence_fedavg_noniid, fl_with_noniid} but also degrades many federated learning algorithms' performances \citep{measuring_the_effects_of_noniid, FL_on_noniid_silos}. By resolving the data heterogeneity problem, learning becomes more robust against partial participation \citep{FedAvg, On_the_convergence_fedavg_noniid}, and the communication cost is also reduced by faster convergence \citep{FedMA, FedMix}.

\begin{figure*}[t!]
    \centering
    \hfill
    \begin{subfigure}[b]{0.315\textwidth}  
        \centering 
        \includegraphics[width=\textwidth]{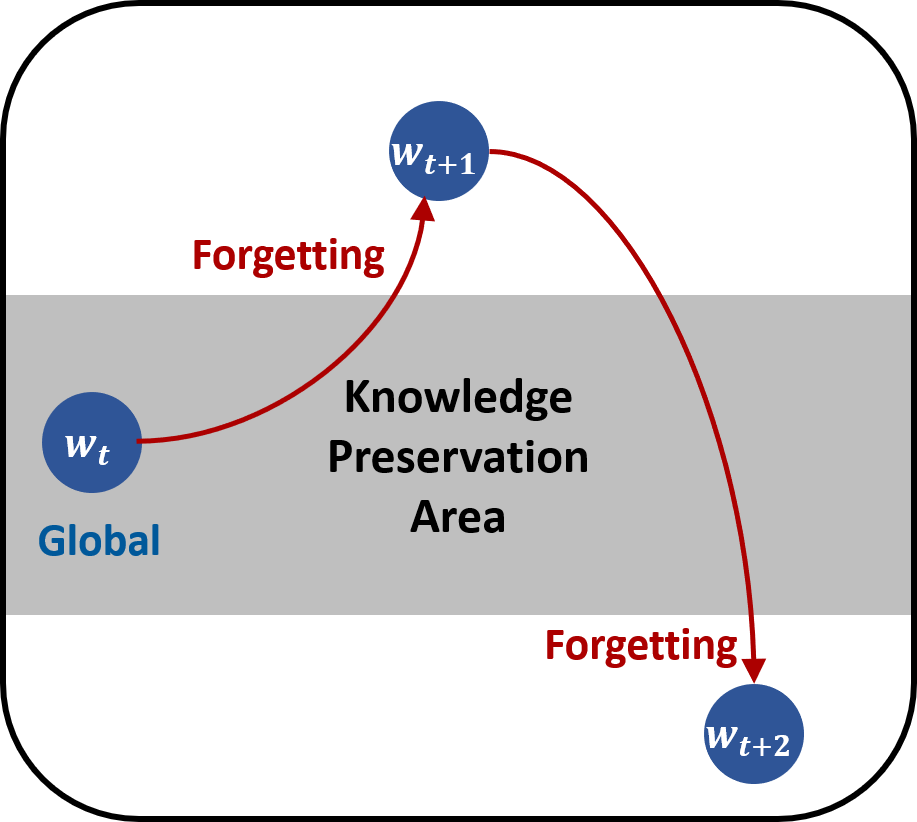}
        \caption[]%
        {{\small Continual Learning}}    
    \label{fig:CL_concept}
    \end{subfigure}
    \hfill
    \begin{subfigure}[b]{0.315\textwidth}  
        \centering 
        \includegraphics[width=\textwidth]{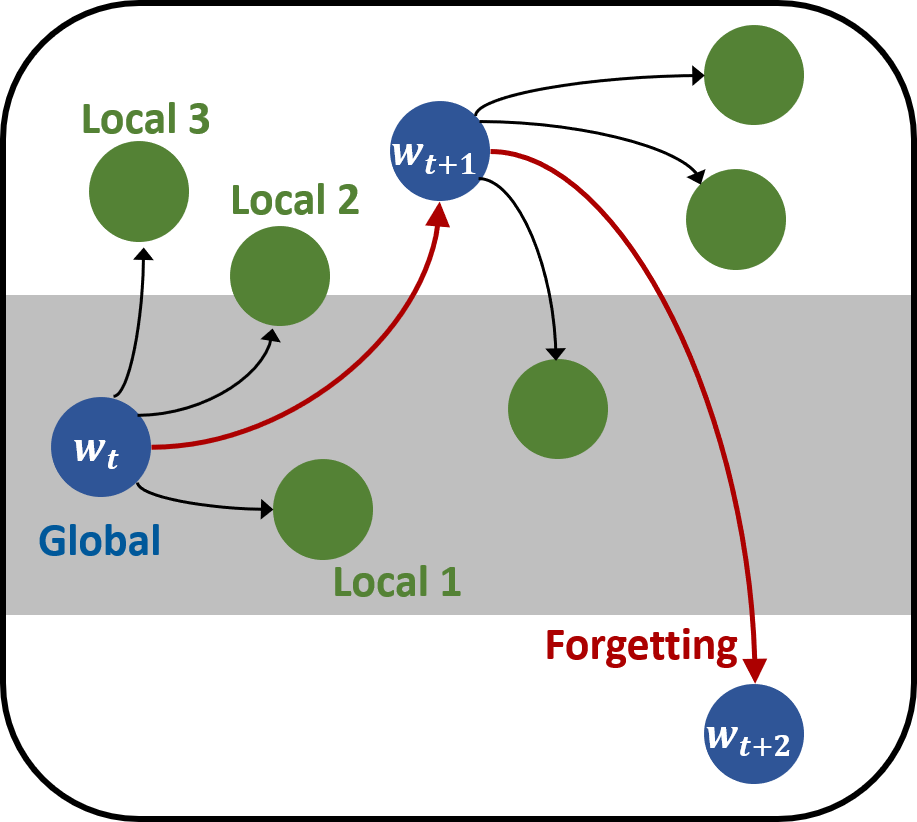}
        \caption[]%
        {{\small Federated Learning (FedAvg)}}    
    \label{fig:FL_concept}
    \end{subfigure}
    \hfill
    \begin{subfigure}[b]{0.315\textwidth}  
        \centering 
        \includegraphics[width=\textwidth]{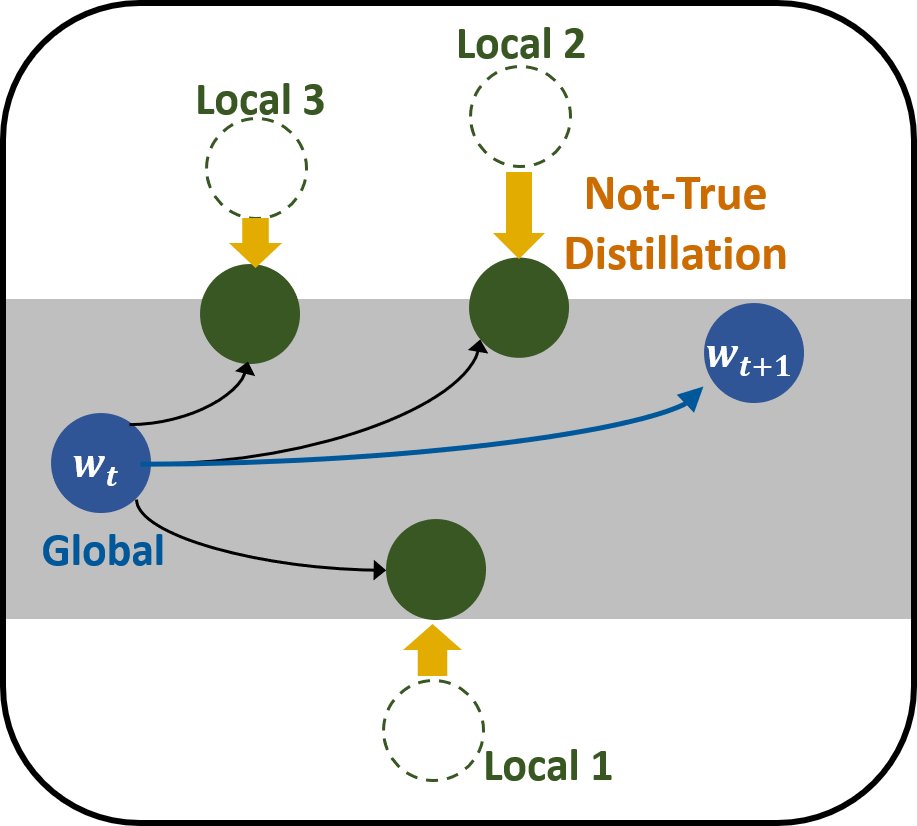}
        \caption[]%
        {{\small Federated Learning (FedNTD)}}
    \label{fig:NTD_concept}
    \end{subfigure}
    \caption[]
    {\small{An overview of forgetting in learning scenarios. As catastrophic forgetting in (a) Continual Learning, (b) Federated Learning also experiences forgetting. However, (c) FedNTD prevents forgetting by preserving global knowledge during local training.}}
    \label{fig:forgetting_concept}
\vspace{-9pt}
\end{figure*}

Interestingly, continual learning \citep{cl_original1, cl_original2} faces a similar challenge. In continual learning, a learner model is continuously updated on a sequence of tasks, with an objective to perform well on whole tasks. Unfortunately, owing to the heterogeneous data distribution of each task, learning on the task sequence often results in \textit{catastrophic forgetting} \citep{cl_forgetting1, cl_forgetting2}, whereby fitting on a new task interferes with the parameters important for previous tasks. As a consequence, the model parameters drift away from the area where the previous knowledge is desirably preserved (\autoref{fig:CL_concept}).

Our first conjecture is such forgetting also exists in federated learning. While the server aggregates local models, the distribution where they are trained may be largely different from those of previous rounds. As a result, the global model faces the distributional shifts at each round, which may cause the forgetting as in continual learning (\autoref{fig:FL_concept}). To empirically verify this analogy, we examine the global model's prediction consistency. More specifically, we measure its class-wise accuracy while the communication rounds proceed.

The observations verify our conjecture: the global model's prediction is highly inconsistent across communication rounds, significantly reducing the performance of predicting some classes that the previous model originally predicted well. We dig deeper to analyze how averaging the locally updated parameters induces such forgetting and confirm that it occurs in local training: the global knowledge, corresponding to the region outside of local distribution, is prone to be forgotten. As merely averaging the local models cannot recover it, the global model struggles to preserve previous knowledge.

Based on our findings, we hypothesize that mitigating the issue of forgetting can relieve data heterogeneity (\autoref{fig:NTD_concept}). To this end, we propose a novel algorithm Federated Not-True Distillation (FedNTD). FedNTD utilizes the global model's prediction on locally available data, but only for the not-true classes. We demonstrate the effect of FedNTD on preserving global knowledge outside of a local distribution and its benefits on federated learning. Experimental results show that FedNTD achieves state-of-the-art performance in various setups.

To summarize, our contributions as follows:
\vspace{-2pt}
\begin{itemize}
    \item We present a systematic study on forgetting in federated learning. The global knowledge outside of the local distribution is prone to be forgotten and is closely related to the data heterogeneity issue \textbf{(Section 2)}.
    \vspace{-2pt}
    \item We propose a simple yet effective algorithm, FedNTD, to prevent forgetting. Unlike prior works, FedNTD neither compromises data privacy nor incurs additional communication burdens. We validate the efficacy of FedNTD on various setups and show that it consistently achieves state-of-the-art performance \textbf{(Section 3, Section 4)}.
    \vspace{-2pt}
    \item We analyze how FedNTD benefits federated learning. The knowledge preservation by FedNTD improves weight alignment and weight divergence after local training \textbf{(Section 5)}.
    \vspace{-2pt}
\end{itemize}

\subsection{Preliminaries}
\vspace{-6pt}
\paragraph{Federated Learning} We aim to train an image classification model in a federated learning system that consists of $K$ clients and a central server. Each client $k$ has a local dataset $\mathcal{D}^k$, where the whole dataset $\mathcal{D} = \bigcup_{k\in [K]}\mathcal{D}^k$. At each communication round $t$, the server distributes the current global model parameters $w^{(t-1)}$ to sampled local clients $K^{(t)}$. Starting from $w^{(t-1)}$, each client $k \in K^{(t)}$ updates the model parameters $w_k^{(t)}$ using its local datasets $\mathcal{D}^{k}$ with the following objective:
\vspace{-2pt}
\begin{equation}
    w_k^{(t)} = \underset{w}{\text{argmin}}\:\:\mathbb{E}_{(x,y) \sim \mathcal{D}^k} [\mathcal{L}(w;w^{(t-1)},x,y)]\,.
\end{equation}
\vspace{-2pt}
where $\mathcal{L}$ is the loss function. At the end of round $t$, the sampled clients upload the locally updated parameters back to the server and aggregate by parameter averaging as $w^{(t)}$ as follows:
\vspace{-2pt}
\begin{equation}
    w^{(t)} = \sum_{k \in K^{(t)}} \frac{|\mathcal{D}^{k}|}{\sum_{k' \in K^{(t)}}|\mathcal{D}^{k'}|}\ w_k^{(t)}.
\end{equation}
\vspace{-12pt}

\paragraph{Knowledge Distillation} Given a teacher model $T$ and a student model $S$, knowledge distillation \citep{knowledge_distillation} matches their softened probability $q_{\mathcal{\tau}}^{T}$ and $q_{\mathcal{\tau}}^{S}$ using temperature $\tau$. The $c$-th value of the $q_{\tau}$ can be described as $q_{\tau}(c) = \frac{ {\exp({z}_{c}/\tau)}}{{\sum_{i}\exp({z}_{i}/\tau)}}$, where $z_c$ is the $c$-th value of logits vector $z$ and $\mathcal{C}$ is the number of classes. Given a sample $x$, the student model $S$ is learned by a linear combination of cross-entropy loss $\mathcal{L}_{\textup{CE}}$ for one-hot label $\mathds{1}_y$ and Kullback-Leibler divergence loss $\mathcal{L}_{\textup{KL}}$ using a hyperparameter $\beta$:
\begin{equation}
    \mathcal{L} = (1-\beta)\mathcal{L}_{\textup{CE}}(q, \mathds{1}_y) + \beta {\tau}^2\mathcal{L}_{\textup{KL}}(q_{\mathcal{\tau}}^{S}, q_{\mathcal{\tau}}^{T})
\end{equation}
\vspace{-8pt}
\begin{equation}
    \mathcal{L}_{\textup{CE}}(q, \mathds{1}_y) =  -\sum_{c=1}^{\mathcal{C}}\mathds{1}_y(c)\log q(c), \qquad     \mathcal{L}_{\textup{KL}}(q_{\mathcal{\tau}}^{S}, q_{\mathcal{\tau}}^{T}) = -\sum_{c=1}^{\mathcal{C}}{q}_{\tau}^{T}(c)\log\left[\frac{{q}_{\tau}^{S}(c)}{{q}_{\tau}^{T}(c)}\right]
\end{equation}
\vspace{-12pt}
\section{Forgetting in Federated Learning}
\vspace{-6pt}
To understand how the non-IID data affects federated learning, we performed an experimental study on heterogeneous locals. We choose CIFAR-10 \citep{cifar} and a convolutional neural network with four layers as in \citep{FedAvg}. We split the data to 100 clients using Latent Dirichlet Allocation (LDA), assigning the partition of class $c$ samples to clients by $p \sim \text{Dir}(\alpha)$. The heterogeneity level increases as the $\alpha$ decreases. We train the model with FedAvg for 200 communication rounds, and 10 randomly sampled clients are optimized for 5 local epochs at each round. More details are in \autoref{appendix_a_setups}.

\subsection{Global Model Prediction Consistency}
\vspace{-6pt}

\begin{figure*}[ht!]
    \centering
    \includegraphics[width=\textwidth]{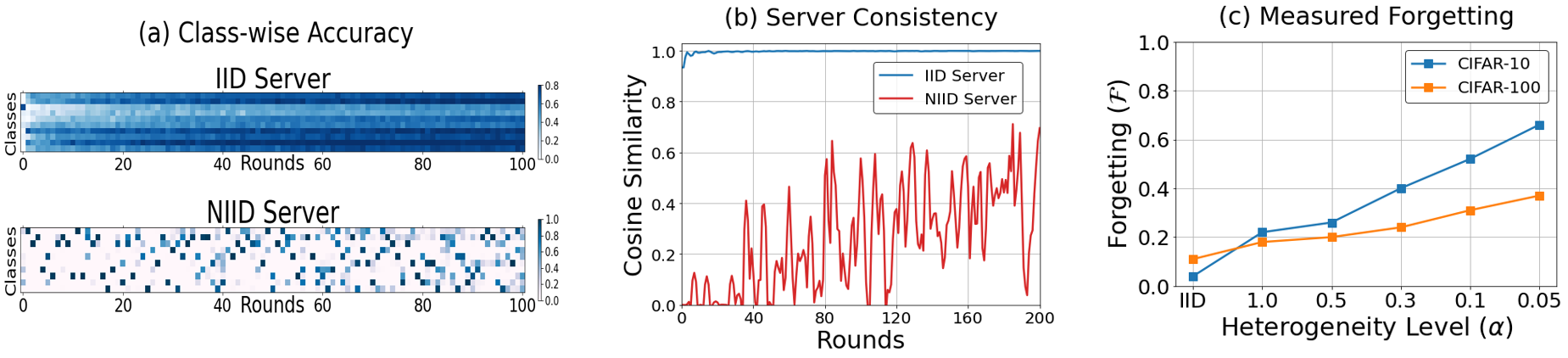}
\vspace{-12pt}
\caption{\small{Forgetting analysis on the global server model. (a): Class-wise test accuracy on CIFAR-10 IID and NIID ($\alpha$=0.1) cases. (b): Cosine similarity of class-wise accuracy vector w.r.t. previous round global model on IID and NIID ($\alpha=0.1$) cases. (c): Forgetting $\mathcal{F}$ by different heterogeneity levels on CIFAR-10 and CIFAR-100.}}
\label{fig2:global_forgetting}
\end{figure*}

To confirm our conjecture on forgetting, we first consider how the global model's prediction varies as the communication rounds proceed. If the data heterogeneity induces forgetting, the prediction after update (i.e., parameter averaging) may be less consistent compared to the previous round. To examine it, we observe the model's class-wise test accuracy at each round, and measure its similarity to the previous round. The results are provided in \autoref{fig2:global_forgetting}\textcolor{red}{a} and \autoref{fig2:global_forgetting}\textcolor{red}{b}.

\vspace{-3pt}
\vspace{-3pt}



As expected, while the server model learned from i.i.d. locals (IID Server) predicts each class evenly well at each round, prediction is highly inconsistent in the non-i.i.d. case (NIID Server). In the non-IID case, the test accuracy on some classes which originally predicted well by the previous global model often drops significantly. This implies that forgetting occurs in federated learning.

To measure how the forgetting is related to data heterogeneity, we borrow the idea of \textit{Backward Transfer} (BwT) \cite{RWalk}, a prevalent forgetting measure in continual learning \citep{a-gem, CPR, cl_low_rank_orthogonal, OGD}, as follows:

\vspace{-10pt}
\begin{equation}
\label{eq:forgetting_measure}
    \mathcal{F} = \frac{1}{\mathcal{C}}\sum_{c=1}^{\mathcal{C}} \max_{t \in \{1, \dots \mathcal{T}-1 \}}(\mathcal{A}_{c}^{(t)} - \mathcal{A}_c^{(\mathcal{T})})
\end{equation}
\vspace{-5pt}

where $\mathcal{A}_c^{(t)}$ is the accuracy on class $c$ at round $t$. Note that the forgetting measure, $\mathcal{F}$, captures the averaged gap between peak accuracy and the final accuracy for each class at the end of learning. The result on the varying heterogeneity levels is plotted in \autoref{fig2:global_forgetting}\textcolor{red}{c}, showing that the global model suffers from forgetting more severely as the heterogeneity level increases.

\subsection{Knowledge Outside of Local Distribution}
\vspace{-6pt}

We take a closer look at local training to investigate why aggregating the local models induces forgetting. In the continual learning view, a straightforward approach is to observe how fitting on the new distribution degrades the performance of the old distribution. However, in our problem setting, the local clients can have any class. Given that only their portions in the local distribution differ across clients, such strict comparison is intractable. Hence, we formulate \textit{\underline{in}-local distribution $p(\mathcal{D})$} and its \textit{\underline{out}-local distribution $\tilde{p}(\mathcal{D})$} to systematically analyze forgetting in local training.

\vspace{-2pt}
\begin{figure*}[ht!]
    \centering
    \includegraphics[width=0.68\textwidth]{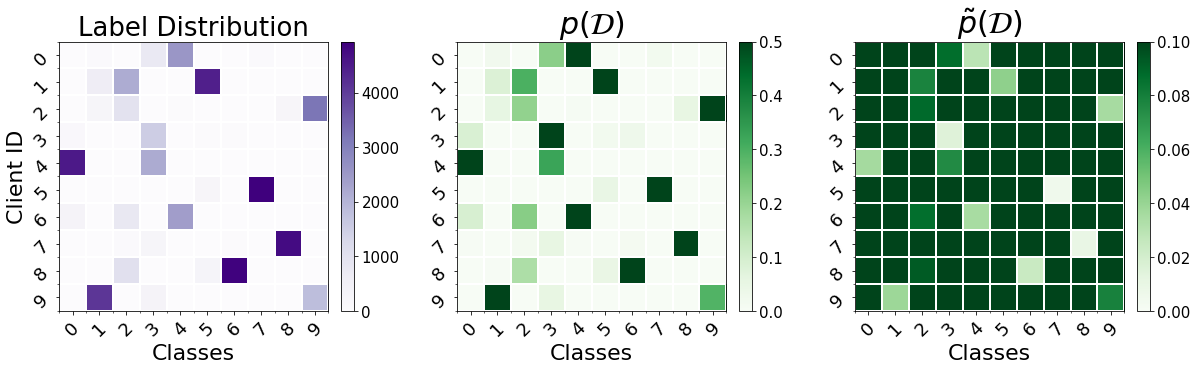}
\caption{\small An example of in-local distribution $p(\mathcal{D})$ and out-local distribution $\tilde{p}(\mathcal{D})$ on CIFAR-10 ($\alpha=0.1$).}
\label{fig3:dist_map}
\end{figure*}

\begin{definition}
Consider a local dataset $\mathcal{D}$ consists of $N$ data points $x_i$ and its label $y_i$ in $\mathcal{C}$-class classification problem. The \textbf{in-local distribution vector $p^k=p(\mathcal{D}^k)$} and its \textbf{out-local distribution vector $\tilde{p}^k=\tilde{p}(\mathcal{D}^k)$} are
\vspace{-10pt}
\begin{align}
p = [p_{1}\,,\ldots,p_{\mathcal{C}}], \quad &\textup{where}\:\: p_c \coloneqq \frac{1}{N}\sum_{i=1}^N \mathbb{I}(y_i = c)\\
\tilde{p} = [\tilde{p}_{1}\,,\ldots,\tilde{p}_{\mathcal{C}}], \quad &\textup{where}\:\: \tilde{p}_{c} \coloneqq \frac{1}{\mathcal{C}-1}( 1 - p_c )
\end{align}
\end{definition}

The underlying idea of out-local distribution $\tilde{p}(\mathcal{D})$ is to assign a higher proportion to the classes with fewer samples in local datasets. Accordingly, it corresponds to the region in the global distribution where the in-local distribution $p(\mathcal{D})$ cannot represent. Note that if $p(\mathcal{D})$ is uniform, $\tilde{p}(\mathcal{D})$ also collapses to uniform, which aligns well intuitively. An example of label distribution for 10 clients and their $p(\mathcal{D})$s and $\tilde{p}(\mathcal{D})$s are provided in \autoref{fig3:dist_map}.

\begin{figure*}[ht!]
    \centering
    \includegraphics[width=0.95\textwidth]{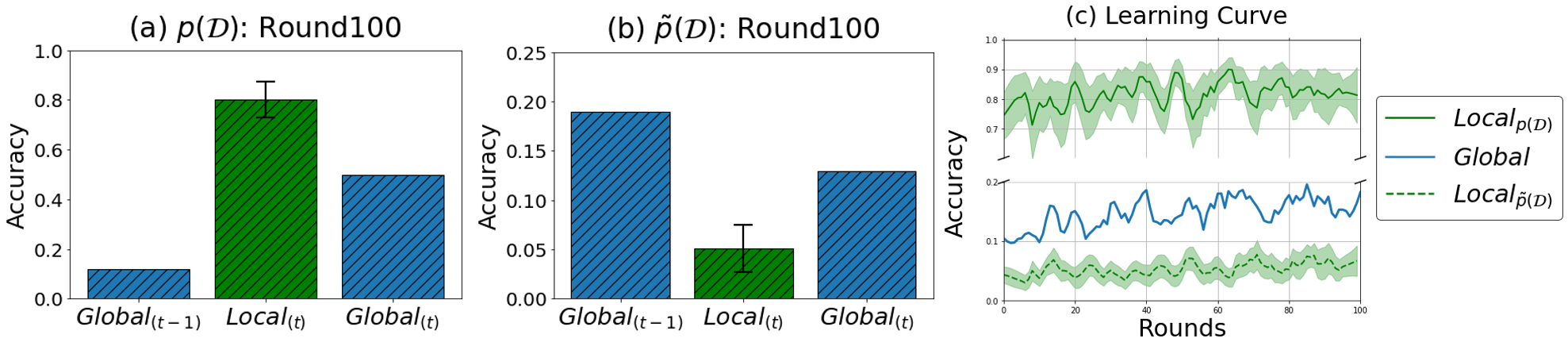}
    \vspace{-3pt}
\caption{\small Accuracy of global model and sampled local models for $p(\mathcal{D})$ and $\tilde{p}(\mathcal{D})$ on CIFAR-10 ($\alpha$=0.1). The error bar stands for the standard deviation of the 10 sampled local clients. In (a) and (b), the global model accuracies for $p(\mathcal{D})$ and $\tilde{p}(\mathcal{D})$ is measured on their joint distributions from 10 sampled clients.}
\label{fig4:local_forgetting}
\end{figure*}

We measure the change of global and local models’ accuracy on $p(\mathcal{D})$ and $\tilde{p}(\mathcal{D})$ during each communication round, as in \autoref{fig4:local_forgetting}. After local training, the local models are well-fitted towards $p(\mathcal{D})$ (\autoref{fig4:local_forgetting}\textcolor{red}{a}), and the aggregated global model also performs well on it. On the other hand, the accuracy on $\tilde{p}(\mathcal{D})$ significantly drops, and the global model accuracy on it also degrades (\autoref{fig4:local_forgetting}\textcolor{red}{b}).

\vspace{-4pt}

To summarize, the knowledge on the out-local distribution $\tilde{p}(\mathcal{D})$ is prone to be forgotten in local training, resulting in the global model's forgetting. Based on our findings, we hypothesize that forgetting could be the stumbling block in federated learning.

\subsection{Forgetting and Local Drift}
\vspace{-6pt}

First empirically observed by \citep{fl_with_noniid}, the deviation of local updates from the desirable global direction has been widely discussed as a major cause of slow and unstable convergence in heterogeneous federated learning \citep{tighter_local_sgd, FedProx, On_the_convergence_fedavg_noniid}. Unfortunately, given the difficulty of analyzing such drift, a common approach is to assume bounded dissimilarity between the local function gradients \citep{SCAFFOLD, On_the_convergence_fedavg_noniid}.

One intriguing property of knowledge preservation on out-local distribution is that it corrects the local gradients towards the global direction. We define the gradient diversity $\Lambda$ to measure the dissimilarity of local gradients and state the effect of knowledge preservation as follows:

\vspace{2pt}
\begin{definition}
For the uniformly weighted $K$ clients, the gradient diversity $\Lambda$ of local functions $f^k$ towards the global function $f=\frac{1}{K} \sum_{k=1}^{K} f^k$ is defined as:
\vspace{-2pt}
\begin{equation}
\Lambda:= \frac{ \frac{1}{K} \sum_{k=1}^{K}\lVert \nabla f^k \rVert^2}{\lVert  \nabla f \rVert^2}\,
\end{equation}
\end{definition}

\vspace{-2pt}
Here, $\Lambda \geq 1$ measures the alignment of gradient direction of the local function $f^k$s w.r.t. the global function $f$. Note that the $\Lambda$ becomes smaller as the directions of local function gradients $\nabla f^k$s become similar---e.g., if the magnitudes of the $\lVert \nabla f^k \rVert^2$s are fixed, the smallest $\Lambda$ is obtained when the direction of $\nabla f^k$s are identical. To understand the effect of preserving knowledge on the out-local distribution $\tilde{p}(\mathcal{D})$, we analyze how the local gradients and their diversity varies by adding gradient signal on $\tilde{p}(\mathcal{D})$ with factor $\beta$ and obtain the following proposition.

\vspace{2pt}
\begin{proposition}
Suppose uniformly weighted $K$ clients with in-local distribution $p^k = [p^k_1, \dots, p^k_\mathcal{C}]$. If we assume the class-wise gradients $g_c$ are orthogonal with uniform magnitude, increasing $\beta \leq \mathcal{C}/2-1$ reduces the gradient diversity $\Lambda$ from the local gradient $\nabla f^k = (p^k + \beta \tilde{p}^k)\cdot g$ with the ratio:
\begin{equation}
\frac{\partial \Lambda}{\partial \beta} \:\: \leq \:\: - \frac{M_{K,\mathcal{C},p}}{(1+\beta)^2} \,.
\end{equation}
\vspace{-1pt}
where $\beta$ stands for the effect of knowledge preservation on the out-local distribution $\tilde{p}^k$. The $M_{K,\mathcal{C},p}>0$ is a constant term consists of $K$, $\mathcal{C}$, and $(p^k)_{k=1}^{K}$\,,
\label{prop0:drift}
\end{proposition}
The proof is given in \autoref{proof_proposition1}.  Note that here we treat $f^k$ as a sum of class-wise losses $\sum_c p^k_c \mathcal{L}_c$, where $\mathcal{L}_c = \mathbb{E}_{x|y=c}[\mathcal{L}(x;w)]$ is the loss on the specific class $c$. When $\beta=0$, there is no regularization to preserve out-distribution knowledge, so the local model only needs to fit on the in-local distribution $p^k$. The above proposition suggests that the preserved knowledge on out-local distribution $\tilde{p}^k$ (i.e., as $\beta$ increases) guides the local gradient directions to be more aligned towards the global gradient, reducing the gradient diversity $\Lambda$. Such forgetting perspective provides the opportunity to handle the data heterogeneity at the model's prediction level.
\vspace{-2pt}
\section{FedNTD: Federated Not-True Distillation}
\vspace{-6pt}

\begin{wrapfigure}{r}{0.5\textwidth}
    \vspace{-18pt}
    \centering
    \includegraphics[width=0.5\textwidth]{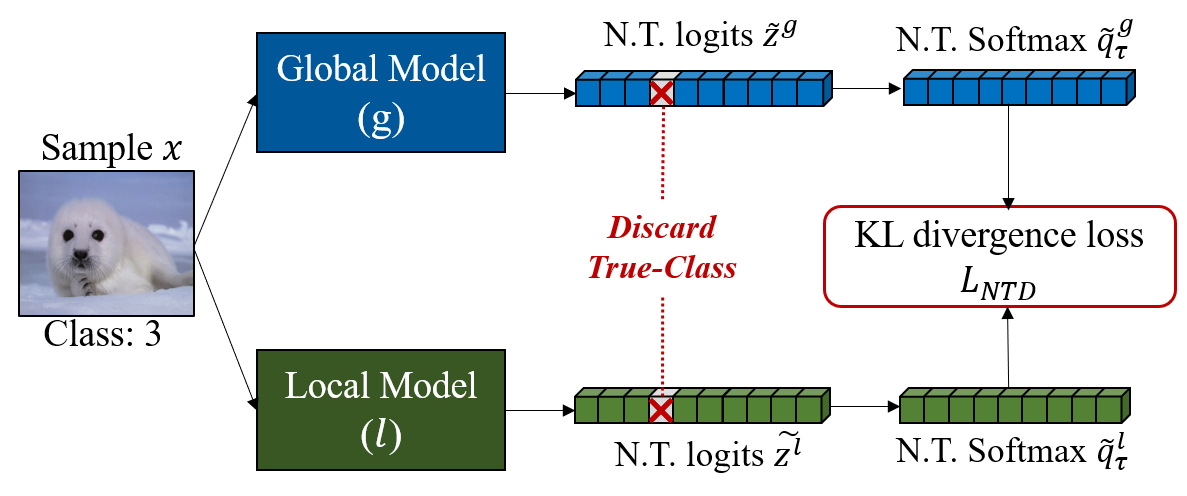}
    \vspace{-15pt}
\caption{\small An overview of Not-True Distillation.The true class (\textit{Class 3}) logits is ignored in the softmax.}
\label{fig:ntd}
\end{wrapfigure}

In this section, we propose Federated Not-True Distillation (FedNTD) and its key features. The core idea of FedNTD is to preserve the global view only for the not-true classes. More specifically, FedNTD conducts local-side distillation by the linearly combined loss function $\mathcal{L}$ between the cross-entropy loss $\mathcal{L}_{\textup{CE}}$ and the not-true distillation loss $\mathcal{L}_{\textup{NTD}}$:
\begin{equation}
\mathcal{L} = \mathcal{L}_{\textup{CE}}(q^{l}, \; \mathds{1}_y) + \beta \cdot \mathcal{L}_{\textup{NTD}}({\tilde{q}_{\tau}^{l}}, \; {\tilde{q}_{\tau}^{g}}).
\label{eq:fedntd}
\end{equation}

\vspace{-4pt}
Here, the hyperparameter $\beta$ stands for the strength of knowledge preservation on the out-local distribution. Then, the not-true distillation loss $\mathcal{L}_{\textup{NTD}}$ is defined as the KL-Divergence loss between the not-true softmax prediction vector ${\tilde{q}_{\tau}}^{l}$ and ${\tilde{q}_{\tau}^{g}}$ as follows:
\begin{equation}
\small
{\mathcal{L}}_{\text{NTD}}({\tilde{q}_{\tau}^{l}},{\tilde{q}_{\tau}^{g}})=-\!\sum_{c=1\,, \underline{\mathbf{c \neq y}}}^{C}{\tilde{q}_{\tau}^{g}}(c)\log\left[\frac{{\tilde{q}_{\tau}^{l}}(c)}{{\tilde{q}_{\tau}^{g}}(c)}\right]\,,
\:\: \text{where}\:\:
\begin{cases}
\normalsize
{\tilde{q}_{\tau}^{l}}(c) =\frac{ \exp{({z}_{c}^{l}/\tau)}}{\sum_{\underline{\mathbf{\tilde{c} \neq y}}}^{C}\exp{({z}_{\tilde{c}}^{l}/\tau)}}
\vspace{8pt}
\\
{\tilde{q}_{\tau}^{g}}(c) =\frac{ \exp{({z}_{c}^{g}/\tau)}}{\sum_{\underline{\mathbf{\tilde{c} \neq y}}}^{C}\exp{({z}_{\tilde{c}}^{g}/\tau)}}
\end{cases} \!\!(\forall c \neq y).
\end{equation}
which take softmax with temperature $\tau$ only for the not-true class logits. \autoref{fig:ntd} illustrates how the not-true distillation works given a sample $x$. Note that ignoring the true-class logits makes gradient signal of $\mathcal{L}_{\textup{NTD}}$ to the true-class as $0$. The detailed algorithm is provided in \textcolor{red}{Algorithm 1}.

\newcommand*{\tikzmk}[1]{\tikz[remember picture,overlay,] \node (#1) {};\ignorespaces}
\newcommand{\algorithmautorefname}{Algorithm}

\newcommand{\boxit}[1]{\tikz[remember picture,overlay]{\node[yshift=3pt,fill=#1,opacity=.25,fit={(A)($(B)+(.95\linewidth,.8\baselineskip)$)}] {};}\ignorespaces}

\begin{algorithm}[ht!]
    \label{algo:fedntd}
    \caption{Federated Not-True Distillation (FedNTD)}
    \textbf{Input:} total rounds ${T}$, local epochs ${E}$, dataset $\mathcal{D}$,
    sampled clients sets ${K}^{(t)}\subset K$ in round $t$, learning rate $\gamma$
    \par
    \smallskip
    \textbf{Initialize} $w^(0)$ for global server weight\par
    \textbf{for} each communication round $t = 1, \cdots, T$ \textbf{do} \par
    \quad Server samples clients $K^{(t)}$ and broadcasts ${\tilde{w}}^{(t)}\leftarrow{w}^{(t)}$ \par
    \quad \textbf{for} each client $k \in K^{(t)}$ \textbf{in parallel do} \par
    \quad \quad \textbf{for} Local Steps $e = 1 \cdots E$ \textbf{do} \par
    \quad \quad \quad \textbf{for} Batches $j = 1 \cdots B$ \textbf{do} \par
    
    \tikzmk{A}
    \quad \quad \quad \quad \small{${\tilde{w}}^{(t)}_k \leftarrow {\tilde{w}}^{(t)}_k - \gamma \nabla_w \mathcal{L}(\,\tilde{w}^{(t)}_k\,;\,[\,\mathcal{D}^k\,]_j\,)$ \qquad \qquad \qquad \textcolor{red}{Using \; [\autoref{eq:fedntd}]}}
    \par
    \tikzmk{B}
    \boxit{black!80}
    \quad \quad \quad \textbf{end for} \par
    \quad \quad \textbf{end for} \par
    \quad \textbf{end for} \par
    \quad Upload ${\tilde{w}}^{t}_k$ to server \par
    \quad \textbf{Server Aggregation :}${w}^{(t+1)} \leftarrow \frac{1}{|K^{(t)}|}\sum_{k \in K^{(t)}} {\tilde{w}}^{(t)}_k$ \par
    \textbf{end for} \par
    \smallskip
    \textbf{Server output :} $w_T$
\end{algorithm}

We now explain how learning to minimize $\mathcal{L}_{\textup{NTD}}$ preserves global knowledge on out-local distribution $\tilde{p}(\mathcal{D})$. Suppose there are $N$ number of data points in the local dataset $\mathcal{D}$. The accumulated Kullback-Leibler divergence loss $\mathcal{L}_{\textup{KL}}$ between $q_{\tau}^{l,i}$, the probability vector for the data $x_i$, and its reference $q_{\tau}^{g,i}$ to be matched for is:
\begin{equation}
\small
\mathcal{L}_{\textup{KL}}=-\frac{1}{N}\sum_{i=1}^{N}\sum_{c=1}^{\mathcal{C}} q_{\tau}^{g,i}(c) \log\left[\frac{q_{\tau}^{l,i}(c)}{q_{\tau}^{g,i}(c)}\right]   \,.
\end{equation}

By splitting the summands for the \textit{true} and \textit{not-true} classes, the term becomes:
\begin{equation}
\small
\mathcal{L}_{\textup{KL}}^{\textup{true}}=-\frac{1}{N}\sum_{i=1}^{N} q_{\tau}^{g,i}(y_i) \log\left[\frac{q_{\tau}^{l,i}(y_i)}{q_{\tau}^{g,i}(y_i)}\right]\,,
\,
\mathcal{L}_{\textup{KL}}^{\textup{not-true}}=-\frac{1}{N}\sum_{i=1}^{N}\sum_{c'\neq y_i}^{\mathcal{C}} q_{\tau}^{g,i}(c') \log\left[\frac{q_{\tau}^{l,i}(c')}{q_{\tau}^{g,i}(c')}\right] \,.
\label{eq:nt_mse}
\end{equation}

\vspace{1pt}
\begin{proposition}
Consider the in-local distribution $p(\mathcal{D}) = [p_1 \dots p_{\mathcal{C}}]$ such that $p_c = \frac{|\mathcal{S}_c|}{N}$ and its  out-local distribution $\tilde{p}(\mathcal{D}) = [\tilde{p}_1, \dots \tilde{p}_{\mathcal{C}}]$, where $\mathcal{S}_c$ is the set of indices satisfying $y_i=c$. Then the $\mathcal{L}_{\textup{KL}}^{\textup{true}}$ and $\mathcal{L}_{\textup{KL}}^{\textup{not-true}}$ each are equivalent to the weighted sum on $p(\mathcal{D})$ and $\tilde{p}(\mathcal{D})$ as
{\small
\begin{align}
\mathcal{L}_{\textup{KL}}^{\textup{true}}= \nsum[1.8]_{c=1}^\mathcal{C} \: &\textcolor{red!70!black}{\boldsymbol{p_c}}  \: \scalebox{1.35}{$\mathbb{E}_{i\in\mathcal{S}_{c}}$}\!\left[- {q_{\tau}^{g,i}(c) \log\left[\frac{q_{\tau}^{l,i}(c)}{q_{\tau}^{g,i}(c)}\right] } \right]\nonumber\\[-8pt]
&\mbox{}\\[-8pt]
\frac{\mathcal{L}_{\textup{KL}}^{\textup{not-true}}}{\mathcal{C}-1}=\nsum[1.8]_{c=1}^\mathcal{C} \: &\textcolor{red!70!black}{\tilde{\boldsymbol{p_c}}}  \: \scalebox{1.35}{$\mathbb{E}_{i\notin\mathcal{S}_{c}}$}\!\left[- {q_{\tau}^{g,i}(c) \log\left[\frac{q_{\tau}^{l,i}(c)}{q_{\tau}^{g,i}(c)}\right] } \right]\nonumber
\end{align}}
\label{prop1:ntd}
\end{proposition}
\vspace{-2pt}
With a minor amount of calculation from \autoref{eq:nt_mse}, we derive the above proposition. The derivation is provided in \autoref{proof_proposition2}. The proposition suggests that matching the true-class and the not-true class logits collapses to the loss on the in-local distribution $p(\mathcal{D})$ and the out-local distribution $\tilde{p}(\mathcal{D})$. 

In the loss function of our FedNTD (\autoref{eq:fedntd}), we attain the new knowledge on the in-local distribution by following the true-class signals from the labeled data in local datasets using the $\mathcal{L}_{\textup{CE}}$. In the meanwhile, we preserve the previous knowledge on the out-local distribution by following the global model's perspective, corresponding to the not-true class signals, using the $\mathcal{L}_{\textup{NTD}}$. Here, the hyperparameter $\beta$ controls the trade-off between learning on the new knowledge and preserving previous knowledge. This resembles to the \textit{stability-plasticity dilemma} \citep{stability-plasticity} in continual learning, where the learning methods must balance retaining knowledge from previous tasks while learning new knowledge for the current task \citep{class_incremental_survey}.
\section{Experiment}
\vspace{-4pt}
\subsection{Experimental Setup}
\vspace{-6pt}

We test our algorithm on MNIST \citep{mnist}, CIFAR-10 \citep{cifar}, CIFAR-100 \citep{cifar}, and CINIC-10 \citep{cinic10}. We distribute the data to 100 clients and randomly sample clients with a ratio of 0.1. For CINIC-10, we use 200 clients, with a sampling ratio of 0.05. We use a momentum SGD with an initial learning rate of 0.1, and the momentum is set as 0.9. The learning rate is decayed with a factor of 0.99 at each round, and a weight decay of 1e-5 is applied. We adopt two different NIID partition strategies: 

\vspace{-2pt}
\begin{itemize}
    \item \textbf{(i) Sharding} \citep{FedAvg}: sort the data by label and divide the data into same-sized shards, and control the heterogeneity by $s$, the number of shards per user. In this strategy only considers the statistical heterogeneity as the dataset size is identical for each client. We set $s$ as MNIST ($s=2$), CIFAR-10 ($s \in \{2, 3, 5, 10\}$), CIFAR-100 ($s=10$), and CINIC-10 ($s=2$).
    \vspace{2pt}
    \item \textbf{(ii) Latent Dirichlet Allocation (LDA)} \citep{CCVR, FedMA}: assigns partition of class $c$ by sampling $p_c \approx \text{Dir}(\alpha)$. In this strategy, both the distribution and dataset size are different for each client. We set $\alpha$ as MNIST ($\alpha=0.1$), CIFAR-10 ($\alpha \in \{0.05, 0.1, 0.3, 0.5\}$), CIFAR-100 ($\alpha=0.1$), and CINIC-10 ($\alpha=0.1$)
\end{itemize}
\vspace{-3pt}
More details on model, datasets, hyperparameters, and partition strategies are provided in \autoref{appendix_a_setups}.

\begingroup
\setlength{\tabcolsep}{1.2pt} 
\renewcommand{\arraystretch}{1.135}
\begin{table*}[t!]
\centering
\caption{Accuracy@1 (\%) on MNIST \citep{mnist}, CIFAR-10 \citep{cifar}, CIFAR-100 \citep{cifar}, and CINIC-10 \citep{cinic10}. The values in the parenthesis are forgetting measure $\mathcal{F}$. The arrow (\scriptsize{\textcolor{red}{$\downarrow$}},\; \scriptsize{\textcolor{green!50!black}{$\uparrow$}})\; \normalsize shows the comparison to the FedAvg. The standard deviation of each experiment is provided in \autoref{appendix_table_std}.}
\vspace{-4pt}
\small
\begin{tabular}{l|ccccccc} 
\toprule
\multicolumn{8}{c}{\textbf{NIID Partition Strategy : Sharding}} \\ 
\hline
\multicolumn{1}{c}{\multirow{2}{*}{\textbf{Method}}} & \multirow{2}{*}{\textbf{MNIST}} & \multicolumn{4}{c}{\textbf{CIFAR-10}}             
& \multirow{2}{*}{\textbf{CIFAR-100}} & \multirow{2}{*}{\textbf{CINIC-10}}  \\
\multicolumn{1}{c}{}    &       & $s=2$     & $s=3$     & $s=5$     & $s=10$    &       &  \\ 
\hline\hline
FedAvg  \scriptsize{\cite{FedAvg}} 
& 78.63${}_{(0.20)}\;\;$ 
& 40.14${}_{(0.59)}\;\;$ 
& 51.10${}_{(0.46)}\;\;$ 
& 57.17${}_{(0.37)}\;\;$  
& 64.91${}_{(0.26)}\;\;$ 
& 25.57${}_{(0.49)}\;\;$ 
& 39.64${}_{(0.59)}\;\;$ \\
\hline

FedCurv  \scriptsize{\cite{FedCurv}} 
& 78.56${}_{(0.21)}$ \scriptsize{\textcolor{red}{$\downarrow$}}  
& 44.52${}_{(0.53)}$ \scriptsize{\textcolor{green!50!black}{$\uparrow$}}
& 49.00${}_{(0.47)}$ \scriptsize{\textcolor{red}{$\downarrow$}}  
& 54.61${}_{(0.39)}$ \scriptsize{\textcolor{red}{$\downarrow$}}  
& 62.19${}_{(0.27)}$ \scriptsize{\textcolor{red}{$\downarrow$}}  
& 22.89${}_{(0.49)}$ \scriptsize{\textcolor{red}{$\downarrow$}}  
& 40.45${}_{(0.57)}$ \scriptsize{\textcolor{green!50!black}{$\uparrow$}} \\


FedProx  \scriptsize{\cite{FedProx}} 
& 78.26${}_{(0.21)}$ \scriptsize{\textcolor{red}{$\downarrow$}}  
& 41.48${}_{(0.57)}$ \scriptsize{\textcolor{green!50!black}{$\uparrow$}}
& 51.65${}_{(0.45)}$ \scriptsize{\textcolor{green!50!black}{$\uparrow$}}
& 56.88${}_{(0.37)}$ \scriptsize{\textcolor{red}{$\downarrow$}} 
& 64.65${}_{(0.25)}$ \scriptsize{\textcolor{red}{$\downarrow$}} 
& 25.10${}_{(0.49)}$ \scriptsize{\textcolor{red}{$\downarrow$}}
& 41.47${}_{(0.57)}$ \scriptsize{\textcolor{green!50!black}{$\uparrow$}} \\


FedNova  \scriptsize{\cite{FedNova}}  
& 77.04${}_{(0.21)}$ \scriptsize{\textcolor{red}{$\downarrow$}} 
& 42.62${}_{(0.56)}$ \scriptsize{\textcolor{green!50!black}{$\uparrow$}}
& 52.03${}_{(0.44)}$ \scriptsize{\textcolor{green!50!black}{$\uparrow$}}
& 62.14${}_{(0.30)}$ \scriptsize{\textcolor{green!50!black}{$\uparrow$}}
& 66.97${}_{(0.20)}$ \scriptsize{\textcolor{green!50!black}{$\uparrow$}}
& 26.96${}_{(0.41)}$ \scriptsize{\textcolor{green!50!black}{$\uparrow$}}
& 42.55${}_{(0.56)}$ \scriptsize{\textcolor{green!50!black}{$\uparrow$}} \\

SCAFFOLD \scriptsize{\cite{SCAFFOLD}}  
& 81.05${}_{(0.17)}$ \scriptsize{\textcolor{green!50!black}{$\uparrow$}}
& 44.60${}_{(0.53)}$ \scriptsize{\textcolor{green!50!black}{$\uparrow$}}
& 54.26${}_{(0.39)}$ \scriptsize{\textcolor{green!50!black}{$\uparrow$}} 
& \textbf{65.74${}_{(0.23)}$} \scriptsize{\textcolor{green!50!black}{$\uparrow$}} 
& \textbf{68.97${}_{(0.16)}$}  \scriptsize{\textcolor{green!50!black}{$\uparrow$}}
& 30.82${}_{(0.36)}$ \scriptsize{\textcolor{green!50!black}{$\uparrow$}}
& 42.66${}_{(0.54)}$ \scriptsize{\textcolor{green!50!black}{$\uparrow$}} \\


MOON  \scriptsize{\cite{MOON}} 
& 76.56${}_{(0.23)}$ \scriptsize{\textcolor{red}{$\downarrow$}}  
& 38.51${}_{(0.60)}$ \scriptsize{\textcolor{red}{$\downarrow$}}  
& 50.47${}_{(0.47)}$ \scriptsize{\textcolor{red}{$\downarrow$}}  
& 56.69${}_{(0.39)}$ \scriptsize{\textcolor{red}{$\downarrow$}}  
& 65.30${}_{(0.25)}$ \scriptsize{\textcolor{green!50!black}{$\uparrow$}}
& 25.29${}_{(0.48)}$ \scriptsize{\textcolor{red}{$\downarrow$}}  
& {37.07}${}_{(0.62)}$ \scriptsize{\textcolor{red}{$\downarrow$}}   \\

\hline
\textbf{FedNTD \scriptsize{(Ours)}}
& \textbf{84.44${}_{(0.13)}$} \scriptsize{\textcolor{green!50!black}{$\uparrow$}}    
& \textbf{52.61${}_{(0.43)}$} \scriptsize{\textcolor{green!50!black}{$\uparrow$}}
& \textbf{58.18${}_{(0.34)}$} \scriptsize{\textcolor{green!50!black}{$\uparrow$}}    
& 64.93${}_{(0.23)}$ \scriptsize{\textcolor{green!50!black}{$\uparrow$}} 
& \textbf{68.56}${}_{(0.15)}$ \scriptsize{\textcolor{green!50!black}{$\uparrow$}} 
& \textbf{31.69${}_{(0.32)}$} \scriptsize{\textcolor{green!50!black}{$\uparrow$}}
& \textbf{48.07${}_{(0.48)}$} \scriptsize{\textcolor{green!50!black}{$\uparrow$}}\\
\hline\hline

\multicolumn{8}{c}{\textbf{NIID Partition Strategy : LDA}}  \\ 
\hline
\multicolumn{1}{c}{\multirow{2}{*}{\textbf{Method}}}  & \multirow{2}{*}{\textbf{MNIST}}  & \multicolumn{4}{c}{\textbf{CIFAR-10}}  
& \multirow{2}{*}{\textbf{CIFAR-100}}  & \multirow{2}{*}{\textbf{CINIC-10}}   \\
\multicolumn{1}{c}{} &   & $\alpha=0.05$ & $\alpha=0.1$ & $\alpha=0.3$  & $\alpha=0.5$ & &  \\ 
\hline\hline
FedAvg  \scriptsize{\cite{FedAvg}} 
& 79.73${}_{(0.19)}\;\;$  
& 28.24${}_{(0.71)}\;\;$ 
& 46.49${}_{(0.51)}\;\;$ 
& 57.24${}_{(0.36)}\;\;$  
& 62.53${}_{(0.28)}\;\;$  
& 30.69${}_{(0.32)}\;\;$ 
& 38.14${}_{(0.60)}\;\;$ \\
\hline

FedCurv  \scriptsize{\cite{FedCurv}} 
& 78.72${}_{(0.20)}$ \scriptsize{\textcolor{red}{$\downarrow$}}  
& 33.64${}_{(0.66)}$ \scriptsize{\textcolor{green!50!black}{$\uparrow$}}
& 44.26${}_{(0.53)}$ \scriptsize{\textcolor{red}{$\downarrow$}}  
& 54.93${}_{(0.38)}$ \scriptsize{\textcolor{red}{$\downarrow$}}  
& 59.37${}_{(0.30)}$ \scriptsize{\textcolor{red}{$\downarrow$}}  
& 29.16${}_{(0.32)}$ \scriptsize{\textcolor{red}{$\downarrow$}}  
& 36.69${}_{(0.61)}$ \scriptsize{\textcolor{red}{$\downarrow$}}  \\


FedProx  \scriptsize{\cite{FedProx}} 
& 79.25${}_{(0.19)}$ \scriptsize{\textcolor{red}{$\downarrow$}}  
& 37.19${}_{(0.62)}$ \scriptsize{\textcolor{green!50!black}{$\uparrow$}}
& 47.65${}_{(0.49)}$ \scriptsize{\textcolor{green!50!black}{$\uparrow$}} 
& 57.35${}_{(0.35)}$ \scriptsize{\textcolor{green!50!black}{$\uparrow$}}
& 62.39${}_{(0.27)}$ \scriptsize{\textcolor{red}{$\downarrow$}}  
& 30.60${}_{(0.32)}$ \scriptsize{\textcolor{red}{$\downarrow$}}
& 39.47${}_{(0.58)}$ \scriptsize{\textcolor{green!50!black}{$\uparrow$}} \\


FedNova  \scriptsize{\cite{FedNova}}  
& 60.37${}_{(0.38)}$ \scriptsize{\textcolor{red}{$\downarrow$}}  
& 10.00 \scriptsize{(\textit{Failed})} \scriptsize{\textcolor{red}{$\downarrow$}}
& 28.06${}_{(0.71)}$ \scriptsize{\textcolor{red}{$\downarrow$}} 
& 57.44${}_{(0.35)}$ \scriptsize{\textcolor{green!50!black}{$\uparrow$}}          
& 64.65${}_{(0.23)}$ \scriptsize{\textcolor{green!50!black}{$\uparrow$}} 
& 32.15${}_{(0.28}$ \scriptsize{\textcolor{green!50!black}{$\uparrow$}}
& 30.44${}_{(0.68)}$ \scriptsize{\textcolor{red}{$\downarrow$}} \\

SCAFFOLD \scriptsize{\cite{SCAFFOLD}}  
& 71.57${}_{(0.26)}$ \scriptsize{\textcolor{red}{$\downarrow$}} 
& 10.00 \scriptsize{(\textit{Failed})} \scriptsize{\textcolor{red}{$\downarrow$}}
& 23.12${}_{(0.74)}$ \scriptsize{\textcolor{red}{$\downarrow$}} 
& \textbf{62.01${}_{(0.29)}$}  \scriptsize{\textcolor{green!50!black}{$\uparrow$}}
& \textbf{66.16${}_{(0.19)}$}  \scriptsize{\textcolor{green!50!black}{$\uparrow$}}
& \textbf{33.68}${}_{(0.25)}$ \scriptsize{\textcolor{green!50!black}{$\uparrow$}}
& 28.78${}_{(0.69)}$ \scriptsize{\textcolor{red}{$\downarrow$}} \\


MOON  \scriptsize{\cite{MOON}} 
& 78.95${}_{(0.20)}$ \scriptsize{\textcolor{red}{$\downarrow$}}  
& 28.35${}_{(0.71)}$ \scriptsize{\textcolor{green!50!black}{$\uparrow$}}
& 44.77${}_{(0.53)}$ \scriptsize{\textcolor{red}{$\downarrow$}}  
& 58.38${}_{(0.35)}$ \scriptsize{\textcolor{green!50!black}{$\uparrow$}}
& 63.10${}_{(0.27)}$ \scriptsize{\textcolor{green!50!black}{$\uparrow$}}
& 30.64${}_{(0.32)}$ \scriptsize{\textcolor{red}{$\downarrow$}}  
& 37.92${}_{(0.61)}$ \scriptsize{\textcolor{red}{$\downarrow$}}  \\

\hline
\textbf{FedNTD \scriptsize{(Ours)}}
& \textbf{81.34${}_{(0.17)}$} \scriptsize{\textcolor{green!50!black}{$\uparrow$}} 
& \textbf{40.17${}_{(0.58)}$} \scriptsize{\textcolor{green!50!black}{$\uparrow$}}
& \textbf{54.42${}_{(0.42)}$} \scriptsize{\textcolor{green!50!black}{$\uparrow$}} 
& \textbf{62.42${}_{(0.29)}$} \scriptsize{\textcolor{green!50!black}{$\uparrow$}} 
& \textbf{66.12${}_{(0.19)}$} \scriptsize{\textcolor{green!50!black}{$\uparrow$}} 
& 32.37${}_{(0.26)}$ \scriptsize{\textcolor{green!50!black}{$\uparrow$}}
& \textbf{46.24${}_{(0.50)}$} \scriptsize{\textcolor{green!50!black}{$\uparrow$}}\\
\bottomrule
\end{tabular}
\label{tab:main_forgetting}
\vspace{-6pt}
\end{table*}
\endgroup

\vspace{-3pt}
\subsection{Performance on Data Heterogeneity}
\vspace{-6pt}
We compare our FedNTD with various existing works, with results shown in \autoref{tab:main_forgetting}. As reported in \citep{FL_on_noniid_silos}, even the state-of-the-art methods perform well only in specific setups, and their performance often deteriorates below FedAvg. However, our FedNTD consistently outperforms the baselines on all setups, achieving state-of-the-art results in most cases.

For each experiment in \autoref{tab:main_forgetting}, we also report the forgetting $\mathcal{F}$ in the parenthesis along with the accuracy. Note that the smaller $\mathcal{F}$ value indicates the global model less forgets the previous knowledge. We find that the performance in federated learning is closely related to forgetting, improving as the forgetting reduces. \textit{We believe that the gain from the prior works is actually from the forgetting prevention in their own ways.}

We emphasize that the prior works to learn from heterogeneous locals often require \textit{statefulness} (i.e., clients should be repeatedly sampled with identification) \citep{FedNova, MOON}, \textit{additional communication cost} \citep{SCAFFOLD}, or \textit{auxiliary data} \citep{fed_ensemble_distill}. However, our FedNTD neither compromise any potential privacy issue nor requires additional communication burden. A brief comparison is provided in \autoref{appendix_prior_works}.

We further conduct experiments on the effect of local epochs, client sampling ratio, and model architecture is in \autoref{appendix_additional}, the advantage of not-true distillation over knowledge distillation in \autoref{appendix_comparision_kd}, and the effect of hyperparameters of FedNTD in \autoref{appendix_ntd_tau_beta}. In the next section, we analyze how the knowledge preservation of FedNTD benefits on the federated learning.
\section{Knowledge preservation of FedNTD}
\vspace{-6pt}

\begin{wrapfigure}{r}{0.557\textwidth}
\vspace{-10pt}
    \centering
    \includegraphics[width=0.557\textwidth]{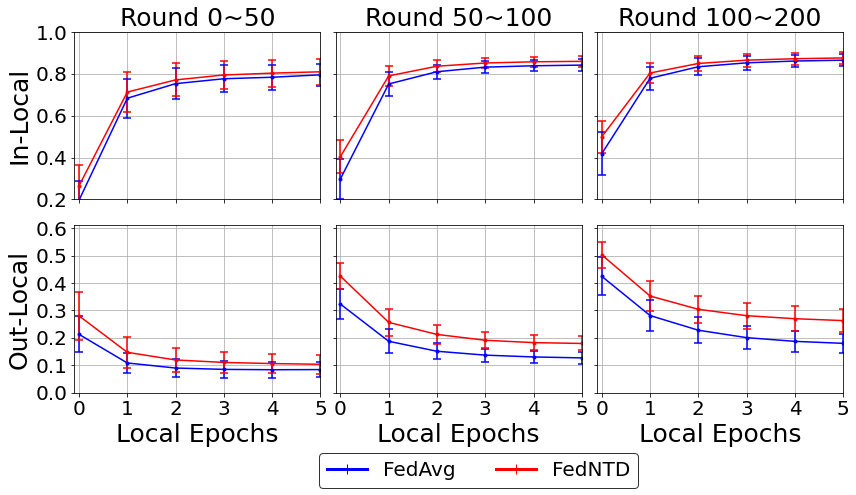}
    \vspace{-10pt}
\caption{\small Accuracy on CIFAR-10 (s=2) in \textit{Local Training}. The error bars stand for the standard deviation on clients.}
\vspace{-7pt}
\label{fig:evidence2}
\end{wrapfigure}

\begin{figure*}[ht!]
    \centering
    \includegraphics[width=0.99\textwidth, height=165pt]{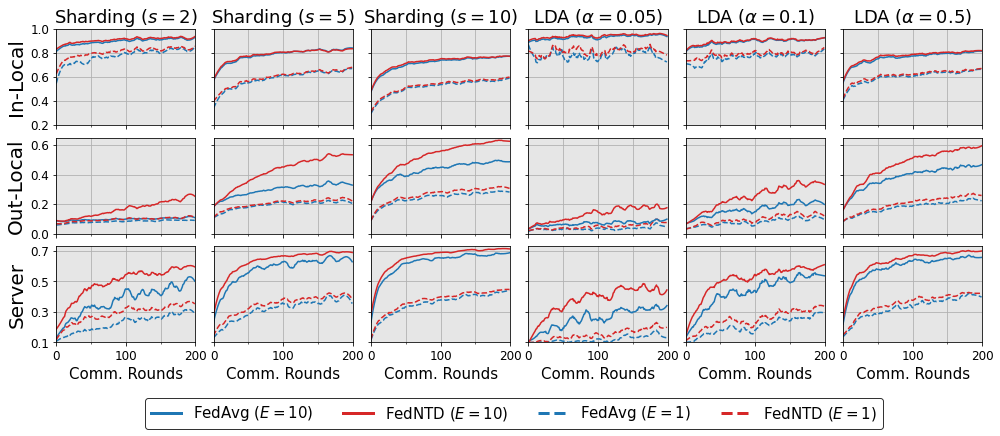}
\caption{Learning curves of FedAvg \citep{FedAvg} (\textit{\textcolor{blue}{blue line}}) and our FedNTD (\textit{\textcolor{red}{red line}}) on CIFAR-10 with various heterogeneity setups for local epochs $E \in \{1, 10\}$. ($\text{1}^{\text{st}}$ and $\text{2}^{\text{nd}}$ row) : Local test accuracy on in-local distribution $p(x)$ and out-local distribution $\tilde{p}(x)$. ($\text{3}^{\text{rd}}$ row): Global server test accuracy.}
\label{fig:evidence1}
\end{figure*}

In \autoref{fig:evidence1}, we present the test accuracy on diverse heterogeneity levels.  Although both FedAvg and FedNTD show little change in local accuracy on in-local distribution $p(x)$, FedNTD significantly improves the local accuracy on out-local distribution $\tilde{p}(x)$, which implies it prevents forgetting. Along with it, the test accuracy of the global model also substantially improves. These gaps are enlarged when the number of local epochs increases, where the local models much deviate from the global model. The accuracy curves during local training are in \autoref{fig:evidence2}. It shows that fitting on the local distribution rapidly leads to forgetting on out-local distribution, But FedNTD effectively relieves this tendency without hurting the learning ability towards the in-local distribution. 

\vspace{-2pt}

Our interest is how the FedNTD's knowledge preservation on out-local distribution benefits on the federated learning, despite little change of its performance on in-local distribution. To figure it out, we analyze the local models in FedNTD after local training, and suggest two main reasons:
\vspace{-2pt}
\begin{itemize}
    \item \textit{\textbf{Weight Alignment}: How much the semantics of each weight is preserved?}
    \item \textit{\textbf{Weight Divergence}: How far the local weights drift from the global model?}
\end{itemize}

\subsection{Weight Alignment}
\vspace{-6pt}

In recent studies, it has been suggested that there is a mismatch of semantic information encoded in the weight parameters across local models, even for the same coordinates (i.e., same position) \citep{PFNM, FedMA, Fed2}. As the current aggregation scheme averages weights of the identical coordinates, matching the semantic alignment across local models plays an important role in global convergence.

To analyze the semantic alignment of each parameter, we identify the individual neuron's class preference by which class has the largest activation output on average. We then measure the alignment for a layer between two different models as the proportion of neurons that the class preference is matched for each other. The result is provided in \autoref{tab:weight_alignment}.

While FedAvg and FedNTD show little difference in the IID case, FedNTD significantly enhances the alignment in the NIID cases. The visualized results are in \autoref{appendix_feature_alignment}, with more details on how the alignment is measured. We further analyze the change of feature after local training using a unit hypersphere in \autoref{appendix_hypersphere} and T-SNE in \autoref{appendix_tsne}.




\vspace{-6pt}
\begingroup
\setlength{\tabcolsep}{5pt} 
\renewcommand{\arraystretch}{1.1}
\begin{table}[ht!]
\caption{\small Alignment of last two fc-layers for Distributed Global (DG), 10 Locals (L), and Aggregated Global (AG) models on CIFAR-10 datasets for IID and NIID (Sharding $s=2$, LDA $\alpha=0.05$) at round 200.}
\vspace{3pt}
\small
\centering
\begin{tabular}{c|c|cc|cc|cc} 
\toprule
\multicolumn{1}{c}{\multirow{2}{*}{\textbf{Layer}}} & \multicolumn{1}{c}{\multirow{2}{*}{\textbf{Alignment}}} 
& \multicolumn{2}{c}{\textbf{IID}}  
& \multicolumn{2}{c}{\textbf{NIID ($s=2$)}}  
& \multicolumn{2}{c}{\textbf{NIID ($\alpha=0.05$)}}\\
\multicolumn{1}{c}{}    & \multicolumn{1}{c}{}   & FedAvg   & \multicolumn{1}{c}{FedNTD} & FedAvg & \multicolumn{1}{c}{FedNTD} & FedAvg & FedNTD \\ 
\hline\hline
\multirow{2}{*}{\begin{tabular}[c]{@{}c@{}}Linear\_1 \\\footnotesize{(dim: 512)}\end{tabular}}  
& $W_{G}^{(t-1)}$ vs. $W_L^{(t)}$          
& 0.679          & 0.668   & 0.635  & \textbf{0.703}     & 0.597   & \textbf{0.756}      \\
& $W_{G}^{(t-1)}$ vs. $W_G^{(t)}$         
& 0.850          & 0.830   & 0.787  & \textbf{0.871}     & 0.670   & \textbf{0.856}      \\ 
\hline
\multirow{2}{*}{\begin{tabular}[c]{@{}c@{}}Linear\_1 \\\footnotesize{(dim: 128)}\end{tabular}}  
& $W_{G}^{(t-1)}$ vs. $W_L^{(t)}$          
& 0.771          & 0.765   & 0.488  & \textbf{0.552}     & 0.512   & \textbf{0.730}  \\
& $W_{G}^{(t-1)}$ vs. $W_G^{(t)}$         
& 0.898          & 0.906   & 0.609  & \textbf{0.836}     & 0.586   & \textbf{0.859}     \\
\bottomrule
\end{tabular}
\label{tab:weight_alignment}
\end{table}
\endgroup

\subsection{Weight Divergence}
\vspace{-4pt}

The knowledge preservation by FedNTD leads the global model to predict each class more evenly. Here we describe how the global model with even prediction performance stabilizes the weight divergence. Consider a model fitted on a specific original distribution, and now it is trained on a new distribution. Then the weight distance between the original model and fitted model increases as the distance between the original distribution and new distribution grows. 

\begin{figure}[ht!]  
    \centering
    \includegraphics[width=\textwidth]{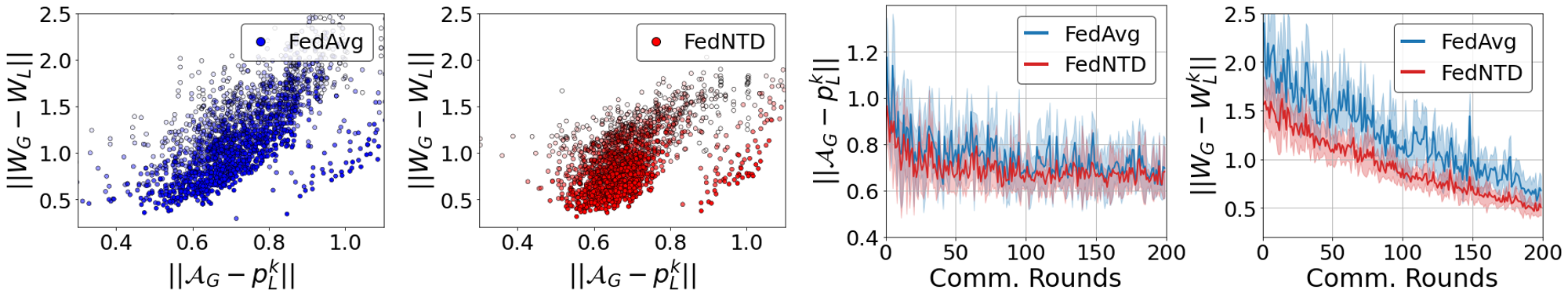}
    \vspace{-13pt}
\caption{\small Distances for weights and distributions on CIFAR10 (s=2). (a), (b): The relationship between two distances. The opacity is higher for later rounds. (c), (d): The measured distances for 200 rounds.}
\label{fig:evidence3}
\end{figure}

We argue that if the distance between the global model's underlying distribution and local distributions is small, the moved distance between the global model and local models also becomes close. If we assume the local distributions are generated arbitrarily, the most robust choice for the global model's underlying distribution is a uniform distribution. We formally rephrase our argument as the follows: 

\vspace{3pt}
\begin{proposition}
Let $\mathbb{P}:\Delta_{\mathcal{C}}\rightarrow\mathbb{R}_{\geq 0}$ be the probability measure for the client's local distribution and $\Pi$ be the set of measure in the hypothesis. Assume that the class-wise loss $\mathcal{L}_c(w) = \mathbb{E}_{x|y=c}[\mathcal{L}(x;w)]$ is $\lambda$-smooth and $w_c$ be a optimum of $\mathcal{L}_c$. Then the loss $\mathcal{L}$ of the client with distribution $p$ becomes:
\begin{equation}
\small
\label{eq:upper_bound}
\mathcal{L}(w) = \sum_c p_c \mathcal{L}_c(w) \leq \sum_c p_c \mathcal{L}_c(w_c) + \frac{1}{2}\lambda \sum_c p_c  \lVert w - w_c \rVert^2\,.
\end{equation}

\vspace{-4pt}
Then for arbitrary $\mathbb{P}\in\Pi$, the uniform distribution attains the minimax value:
\begin{equation}
\label{eqn: cor2}
\text{unif.dist} \in  \underset{p\in \Delta_{\mathcal{C}}}{\textup{argmin}}  \:\underset{\mathbb{P}\in\Pi}{\textup{sup}}\:\mathbb{E}_{p'\sim\mathbb{P}}[\lVert w_{p'} - w_{p} \rVert]\,,\:\: \textup{where}\:\: w_p = \sum_c p_c w_c\,.
\end{equation}
\end{proposition}

\vspace{-3pt}
The proof is provided in \autoref{proof:proposition3}. Although the global model's underlying distribution is unknown, the normalized class-wise accuracy vector is a handy approximation for it as: $\mathcal{A_{\text{G}}} = \frac{1}{A}\cdot[a_1, \dots a_{\mathcal{C}}]$, where $A$ is the global model's test accuracy and $a_c$ is its class-wise accuracy on the class $c$. 

The results in \autoref{fig:evidence3} empirically validate our argument. There is a strong correlation between weight divergence $\|w - w_k\|$ (for global model $w$ and client $k$'s local model $w_k$) and distribution distance $\|\mathcal{A}_G - p_k\|$ (for client $k$'s distribution $p_k$). By providing a better starting point for local training, FedNTD effectively stabilizes the weight divergence.

\section{Related Work}
\vspace{-6pt}

\paragraph{Federated Learning (FL)} is proposed to update a global model while the local data is kept in clients' devices \citep{federated_learning, federated_optimization}. The standard algorithm is FedAvg \citep{FedAvg}, which aggregates trained local models by averaging their parameters. Although its effectiveness has been largely discussed in i.i.d. settings  \citep{local_sgd_better, local_sgd_converges_fast}, many algorithms obtain the sub-optimal when the distributed data is heterogeneous \citep{On_the_convergence_fedavg_noniid, fl_with_noniid}. Until recently, a wide range of variants of FedAvg has been proposed to overcome such a problem. One line of work focuses on \textit{local-side} modification by regularizing the deviation of local models from the global model \citep{FedProx, SCAFFOLD, FedDyn, FedNova}. Another is the \textit{server-side} modification, which improves the efficacy of aggregation of local models in the server \citep{FedMA, fed_ensemble_distill, FedFTG, FedBE}. Our work aims to preserve global knowledge during local training, which belongs to the local-side approach.
\vspace{-4.5pt}

\paragraph{Forgetting View in FL} A pioneer work that considers forgetting in FL is FedCurv \citep{FedCurv}. It regards each local client as a task, and \citep{FedCurv} regulates the change of local parameters to prevent accuracy drop on all other clients. However, it needs to compute and communicate parameter-wise importance across clients, which severely burdens the learning process. On the other hand, we focus on the class-wise forgetting and suggest that not-true logits from local data contain enough knowledge to prevent it. A concurrent work of our study is \citep{FedReg}, which also reports the forgetting issue in local clients by empirically showing the increasing loss of previously learned data after the local training. To prevent forgetting, \citep{FedReg} exploits generated pseudo data. Instead, we focus on the class-wise forgetting and suggest that not-true logits from local data contain enough knowledge to prevent it. The continual learning literature is further discussed in \autoref{appendix_related_work}.
\vspace{-4.5pt}

\paragraph{Knowledge Distillation (KD) in FL} In FL, a typical approach is using KD to make the global model learn from the ensemble of local models \citep{FedBE, fed_ensemble_distill, FedMD, distilled_one_shot}. By leveraging the unlabeled auxiliary data, KD effectively tackles the local drifts by enriching the aggregation. However, such carefully engineered proxy data and may not always be available \citep{FedProx, FedGEN, FedFTG}. Although more recent works generate pseudo-data to extract knowledge by data-free KD \citep{FedGEN, FedFTG}, they require additional heavy computation, and the quality of samples is sensitive to the many hyperparameters involved in the process. On the other hand, as a simple variant of KD, our proposed method surprisingly performs well on heterogeneity scenarios without any additional resource requirements.
\vspace{-2pt}
\section{Conclusion}
\vspace{-6pt}
This study begins from an analogy to continual learning and suggests that forgetting could be a major concern in federated learning. Our observations show that the knowledge outside of local distribution is prone to be forgotten in local training and is closely related to the unstable global convergence. To overcome this issue, we propose a simple yet effective algorithm, FedNTD, which conducts local-side distillation only for the not-true classes to prevent forgetting. FedNTD does not have any additional requirements, unlike previous approaches. We analyze the effect of FedNTD from various perspectives and demonstrate its benefits in federated learning.
\vspace{-4pt}

\paragraph{Broader Impact} We believe that Federated Learning is an important learning paradigm that enables privacy-preserving ML. Our work suggests the forgetting issue and introduces the methods to relive it without compromising data privacy. The insight behind this work may inspire new researches. However, the proposed method maintains the knowledge outside of local distribution in the global model. This implies that if the global model is biased, the trained local model is more prone to have a similar tendency. This should be considered for ML participators.
\vspace{-4pt}

\section*{Acknowledgments}
\vspace{-6pt}
This work was supported by Institute of Information \& communications Technology Planning \& Evaluation (IITP) grant funded by Korea government (MSIT) [No. 2021-0-00907, Development of Adaptive and Lightweight Edge-Collaborative Analysis Technology for Enabling Proactively Immediate Response and Rapid Learning, 90\%] and [No. 2019-0-00075, Artificial Intelligence Graduate School Program (KAIST), 10\%].

\clearpage
\bibliography{main}
\bibliographystyle{plain}
\clearpage
\section*{Checklist}

The checklist follows the references.  Please
read the checklist guidelines carefully for information on how to answer these
questions.  For each question, change the default \answerTODO{} to \answerYes{},
\answerNo{}, or \answerNA{}.  You are strongly encouraged to include a {\bf
justification to your answer}, either by referencing the appropriate section of
your paper or providing a brief inline description.  For example:
\begin{itemize}
  \item Did you include the license to the code and datasets? \answerYes{The datasets are public datasets and the code is MIT license}
\end{itemize}
Please do not modify the questions and only use the provided macros for your
answers.  Note that the Checklist section does not count towards the page
limit.  In your paper, please delete this instructions block and only keep the
Checklist section heading above along with the questions/answers below.

\begin{enumerate}

\item For all authors...
\begin{enumerate}
  \item Do the main claims made in the abstract and introduction accurately reflect the paper's contributions and scope?
    \answerYes{We clarified the scope in both abstract and introduction. The contributions are summarized at the end of Section 1.}
  \item Did you describe the limitations of your work?
    \answerYes{Before reference, we discussed about potential limitations as broad impact}
  \item Did you discuss any potential negative societal impacts of your work?
    \answerYes{Before reference, we discussed as broader impact}
  \item Have you read the ethics review guidelines and ensured that your paper conforms to them?
    \answerYes{}
\end{enumerate}

\item If you are including theoretical results...
\begin{enumerate}
  \item Did you state the full set of assumptions of all theoretical results?
    \answerYes{In all proposition parts. The detailed notations are in Appendix A.}	
\item Did you include complete proofs of all theoretical results?
    \answerYes{Included in the corresponding Appendix sections.}
\end{enumerate}

\item If you ran experiments...
\begin{enumerate}
  \item Did you include the code, data, and instructions needed to reproduce the main experimental results (either in the supplemental material or as a URL)?
    \answerYes{Included in supplemental material}
  \item Did you specify all the training details (e.g., data splits, hyperparameters, how they were chosen)?
    \answerYes{Included in the Appendix.}
	\item Did you report error bars (e.g., with respect to the random seed after running experiments multiple times)?
    \answerYes{We indicated the standard deviation of each result for the main experiments in the corresponding Appendix section.}
	\item Did you include the total amount of compute and the type of resources used (e.g., type of GPUs, internal cluster, or cloud provider)?
    \answerYes{Included in the Appendix.}
\end{enumerate}

\begin{enumerate}
  \item If your work uses existing assets, did you cite the creators?
    \answerYes{}
  \item Did you mention the license of the assets?
    \answerYes{}
  \item Did you include any new assets either in the supplemental material or as a URL?
    \answerYes{}
  \item Did you discuss whether and how consent was obtained from people whose data you're using/curating?
    \answerNA{We used the public benchmarks.}
  \item Did you discuss whether the data you are using/curating contains personally identifiable information or offensive content?
    \answerNA{We only used the public dataset that has been sufficiently considered about the issues.}
\end{enumerate}

\item If you used crowdsourcing or conducted research with human subjects...
\begin{enumerate}
  \item Did you include the full text of instructions given to participants and screenshots, if applicable?
    \answerNA{}
  \item Did you describe any potential participant risks, with links to Institutional Review Board (IRB) approvals, if applicable?
    \answerNA{}
  \item Did you include the estimated hourly wage paid to participants and the total amount spent on participant compensation?
    \answerNA{}
\end{enumerate}
\end{enumerate}

\appendix{\section{Table of Notations}
\begin{table}[h]
\caption{Table of Notations throughout the paper.}
\begin{tabularx}{\textwidth}{p{0.35\textwidth}X}
      \toprule
      {Indices: } \\
      $c\,,c'$ & index for classes ($c\in\{1,...,\mathcal{C}\}=[\mathcal{C}]$)  \\ 
      $i$ & index for data ($i\in\{1,...,N\}=[N]$)\\
      $k\,,k'$ & index for clients ($k\in\{1,...,K\}=[K]\text{ or }\in K^{(t)}$)\\
      $t$ & index for rounds ($t\in\{1,...,\mathcal{T}\}=[\mathcal{T}]$)\\
      $e$ & index for local epochs ($t\in\{1,...,E\}=[E]$)\\
      \midrule
      {Parameters: } \\
      $\alpha$   & Parameter for the Dirichlet Distribution\\
      $s$        & The Number of shards per user\\
      $\beta$    & Hyperparameter for the distillation loss; generally controls the relative weight of divergence loss\\
      $\tau$     & The temperature on the softmax \\
      $\gamma$   & Learning rate\\
      \midrule
      {Data and Weights: } \\
      $\mathcal{D}$    & whole dataset\\
      $\mathcal{D}^k$  & local dataset\\
      $x$   & datum  \\
      $y$   & class label for datum \\
      $w^{(t)}$   & weight of the server model on the round $t$\\
      $w_k^{(t)}$ & weight of the $k$-th client model on the round $t$\\
      $\Vert W_G - W_L \Vert$ &  Collection of $L^1$-norm between server and client models, among all rounds.\\
      \midrule
      {Softmax Probabilities: } \\
      $q_{\mathcal{\tau}}^{T}\,/\,q_{\mathcal{\tau}}^{S}$  & The softened softmax probability of teacher/student model\\
      $q^g\,/\,q^l$  & The softmax probability on the server/client model\\
      ${\tilde{q}_{\tau}}^{g}\,/\,{\tilde{q}_{\tau}^{l}}$ & The softened softmax probability calculated without true-class logit on the server/client model\\
      \midrule
      {Class Distribution on Datasets: } \\
      $p = [p_{1}\,,\ldots,p_{\mathcal{C}}]$  & In-local distribution of dataset \\
      $\tilde{p} = [\tilde{p}_{1}\,,\ldots,\tilde{p}_{\mathcal{C}}]$ &  Out-local distribution of dataset\\
      \midrule
      {Loss Functions: }\\
      $\mathcal{L}_{\textup{CE}}$  & cross-entropy loss\\
      $\mathcal{L}_{\textup{KL}}$  & Kullback-Leibler divergence\\
      $\mathcal{L}_{\textup{NTD}}$ & Proposed Not-True Distillation Loss\\
      \midrule
      {Accuracy and Forgetting Measure: }\\
      $\mathcal{A}_c^{(t)}$  & Accuracy of the server model on the $c$-th class, at round $t$.\\
      $\Vert \mathcal{A}_{G} - p_L^k \Vert$ & Collection of $L^1$-norm between \textit{normalized} global accuracy and data distribution on each client, among all rounds.\\
      $\mathcal{F}$  & Backward Transfer (BwT). For the federated learning situation, we calculate this measure on the server model.\\ 
      \bottomrule
\end{tabularx}
\end{table}


\clearpage
\section{Experimental Setups}
\label{appendix_a_setups}
Here we provide details of our experimental setups. The code is implemented by PyTorch \citep{pytorch} and the overall code structure is based on FedML \citep{fedml} library with some modifications. We use 1 Titan-RTX and 1 RTX 2080Ti GPU card. Multi-GPU training is not conducted in the paper experiments.

\subsection{Model Architecture}
The model architecture used in our experiment is from \citep{FedAvg}, which is composed of two convolutional layers followed by max-pooling layers, and two fully connected layer. A similar architecture is used in \citep{MOON, CCVR}.

\subsection{Datasets}
We mainly used four benchmarks: MNIST \citep{mnist}, CIFAR-10 \citep{cifar}, CIFAR-100 \citep{cifar}, and CINIC-10 \citep{cinic10}. The details about each datasets and setups are described in \autoref{tab:datasets}. We augment the training data using Random Cropping, Horizontal Flipping, and Normalization. For MNIST, CIFAR-10, and CIFAR-100, we add Cutout \citep{cutout} augmentation.

\begin{table}[ht!]
\caption{\small Details datasets setups used in the experiment.}
\centering
\begin{tabular}{ccccc} 
\toprule
\textbf{Datasets}     & \textbf{MNIST} & \textbf{CIFAR-10} & \textbf{CIFAR-100} & \textbf{CINIC-10}  \\ 
\hline
Datasets Classes      & 10             & 10                & 100                & 10                 \\
Datasets Size         & 50,000         & 50,000            & 50,000             & 90,000             \\
Number of Clients     & 100            & 100               & 100                & 200                \\
Client Sampling Ratio & 0.1            & 0.1               & 0.1                & 0.05               \\
Local Epochs (E)      & 3              & 5                 & 5                  & 5                  \\
Batch Size (B)        & 50             & 50                & 50                 & 50                 \\
\bottomrule
\end{tabular}
\label{tab:datasets}
\end{table}

\subsection{Learning Setups}
We use a momentum SGD optimizer with an initial learning rate of 0.01, and the momentum is set as 0.9. The momentum is only used in the local training, which implies the momentum information is not communicated to the server. The learning rate is decayed with a factor of 0.99 at each round, and a weight decay of 0.00001 is applied. In the motivational experiment in Section 3, we fix the learning rate as 0.01. Since we assume a synchronized federated learning scenario, parallel distributed learning is simulated by sequentially training the sampled clients and then aggregating them as a global model.

\subsection{Algorithm Implementation Details}

For the implemented algorithms, we search hyperparameters and choose the best among the candidates. The hyperparameters for each algorithm is in \autoref{tab:algorithm_hyperparameters}.

\begingroup
\setlength{\tabcolsep}{6.0pt} 
\renewcommand{\arraystretch}{1.1}
\begin{table}[ht!]
\caption{\small Algorithm-specific hyperparameters used in the experiment.}
\small
\centering
\begin{tabular}{lll}
\toprule
\textbf{Method}        & \textbf{Hyperparameters} & \textbf{Searched Candidates}  \\ 
\hline\hline
FedAvg \citep{FedAvg}                 
& None                     
& None
\\
FedCurv \citep{FedCurv}               
& $s=500$, $\lambda=1.0$       
& $s \in \{250, 500\}$, $\lambda \in \{0.1, 0.5, 1.0\}$
\\
FedProx \citep{FedProx}
& $\mu = 0.1$
& $\mu \in \{0.1, 0.5, 1.0\}$
\\
FedNova \citep{FedNova}
& None
& None
\\
SCAFFOLD \citep{SCAFFOLD}
& None
& None
\\
MOON \citep{MOON}
& $\mu = 1.0$, $\tau = 0.5$
& $\mu \in \{0.1, 0.5, 1.0\}$, $\tau \in \{0.1, 0.5, 1,0\}$
\\ 
\hline
\textbf{FedNTD (Ours)} & $\beta$=1.0, $\tau=1.0$ & None \\
\bottomrule
\end{tabular}
\label{tab:algorithm_hyperparameters}
\end{table}
\endgroup

\clearpage
\subsection{Non-IID Partition Strategy}

 To widely address the heterogeneous federated learning scenarios, we distribute the data to the local clients with the following two strategies: (1) \textbf{Sharding} and (2) \textbf{Latent Dirichlet Allocation} (LDA).
 
 \paragraph{Sharding} In the Sharding strategy, we sort the data by label and divide it into the same-sized shards without overlapping. In detail, a shard contains $\frac{|D|}{N \times s}$ size of same class samples, where $D$ is the total dataset size, $N$ is the total number of clients, and $s$ is the number of shards per user. Then, we distribute \textit{s} number of shards to each client. $s$ controls the heterogeneity of local data distribution. The heterogeneity level increases as the shard per user $s$ becomes smaller and vice versa. Note that we only test the statistical heterogeneity (skewness of class distribution across the local clients) in the Sharding strategy, and the size of local datasets is identical.
 
 \paragraph{Latent Dirichlet Allocation (LDA)} In LDA strategy, Each client $k$ is allocated with $p_{c,k}$ proportion of the training samples of class $c$, where $p_c \sim \text{Dir}_K(\alpha)$ and $\alpha$ is the concentration parameter controlling the heterogeneity. 
The heterogeneity level increases as the concentration parameter $\alpha$ becomes smaller, and vice versa. Note that both the class distribution and local datasets sizes differ across the local clients in LDA strategy.

\section{Conceptual comparison to prior works}
\label{appendix_prior_works}

The conceptual illustration of federated distillation methods is in \autoref{fig:distill_method_overview}. The existing algorithms either use additional local information (\autoref{fig:distill_a}) or need an auxiliary (or \textit{proxy}) data to conduct distillation. On the other hand, our proposed FedNTD does not have such constraints (\autoref{fig:distill_c}).

\begin{figure*}[ht!]
    \centering
    \hfill
    \begin{subfigure}[b]{0.325\textwidth}  
        \centering 
        \includegraphics[width=\textwidth]{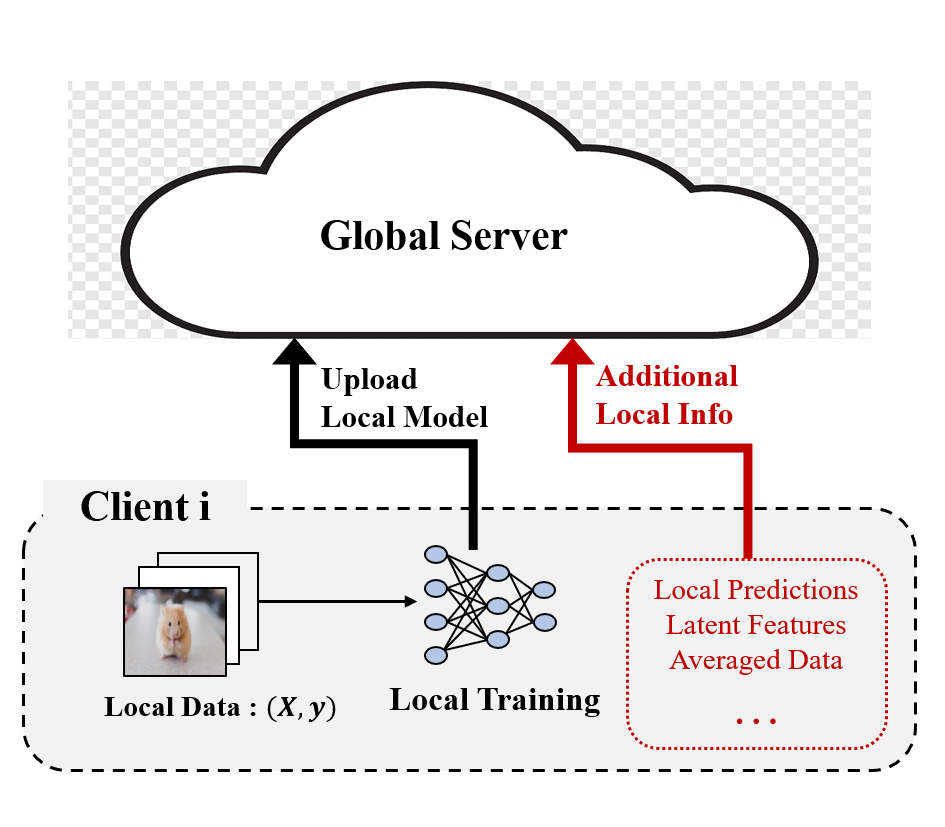}
        \caption[]%
        {{\small Use Additional Local Info}}
        \label{fig:distill_a}
    \end{subfigure}
    \hfill
    \begin{subfigure}[b]{0.325\textwidth}  
        \centering 
        \includegraphics[width=\textwidth]{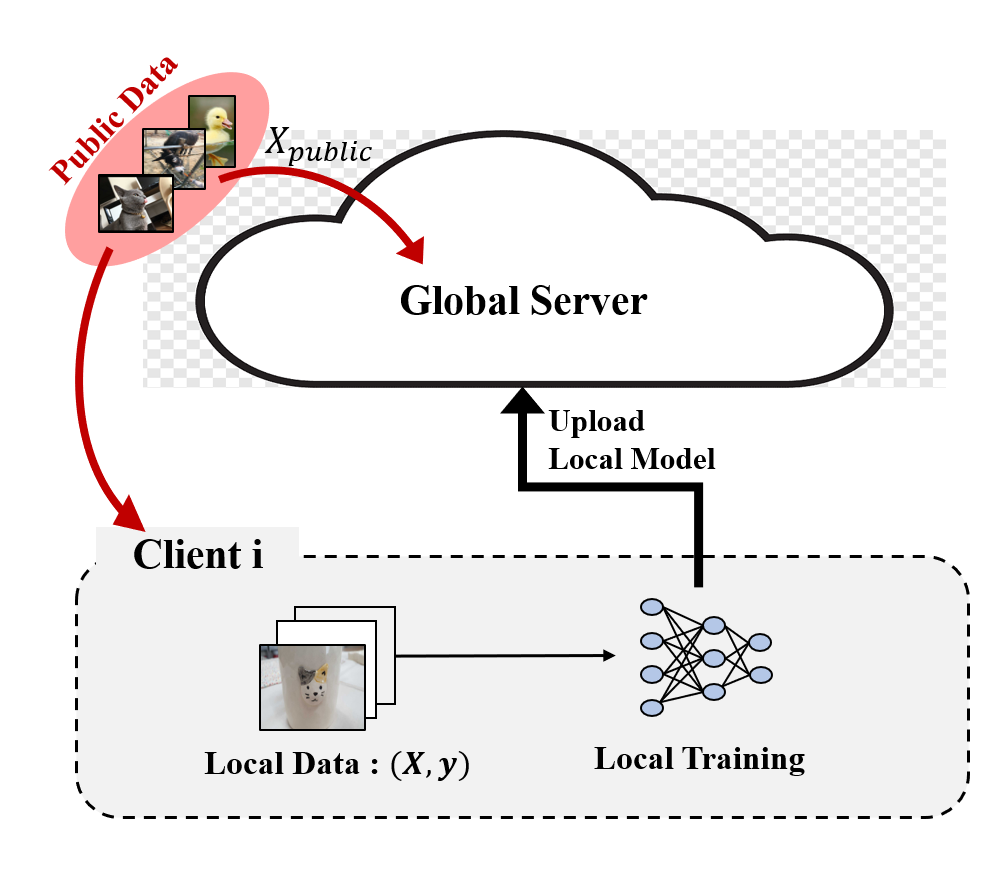}
        \caption[]%
        {{\small Use Auxiliary Datasets}}
        \label{fig:distill_b}
    \end{subfigure}
    \hfill
    \begin{subfigure}[b]{0.325\textwidth}  
        \centering 
        \includegraphics[width=\textwidth]{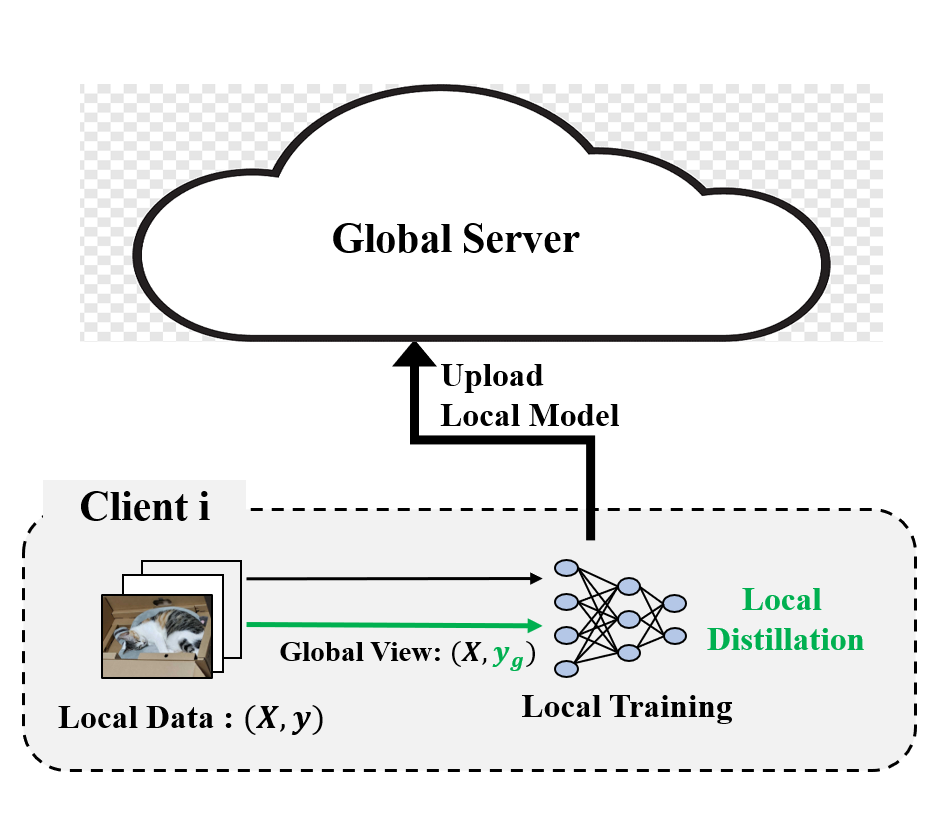}
        \caption[]%
        {{\small FedNTD (ours)}}
        \label{fig:distill_c}
    \end{subfigure}
    \caption[]
    {\small An overview of federated distillation methods.}
\label{fig:distill_method_overview}
\end{figure*}

\begingroup
\newcommand{\xmark}{\textcolor{red!80!black}{\ding{53}}}
\newcommand{\cmark}{\textcolor{green!50!black}{\ding{52}}}
\setlength{\tabcolsep}{6.0pt} 
\renewcommand{\arraystretch}{1.15}
\begin{table}[ht!]
\caption{\small Additional resource requirements compared to FedAvg.}
\footnotesize
\centering
\begin{tabular}{l|ccc}
\toprule
\multicolumn{1}{c}{\multirow{2}{*}{\textbf{Method}}} & \multicolumn{3}{c}{\textbf{No Additional Requirements on:}}  \\
\multicolumn{1}{c}{}    & Statefulness? & Communication Cost? &  Auxiliary Data? \\ 
\hline\hline
FedEnsemble \citep{fed_ensemble_distill}                            
& \cmark        & \cmark         & \xmark  \\
FedBE \citep{FedBE}                                  
& \cmark        & \xmark         & \xmark  \\
MOON \citep{MOON}                                
& \xmark        & \cmark         & \cmark  \\
SCAFFOLD \citep{SCAFFOLD}               
& \xmark        & \xmark         & \cmark  \\
\hline
\textbf{FedNTD (Ours)}  & \cmark        & \cmark         & \cmark \\
\bottomrule
\end{tabular}
\label{tab:additional_requirements}
\end{table}
\endgroup

\clearpage
\section{Related Work: Continual Learning}
\label{appendix_related_work}

Continual Learning (CL) is a learning paradigm that updates a sequence of tasks instead of training on the whole datasets at once  \citep{cl_original1, cl_original2}. In CL, the main challenge is to avoid catastrophic forgetting \citep{catastrophic_forgetting}, whereby training on the new task interferes with the previous tasks. Existing methods try to mitigate this problem by various strategies. In \textit{Parameter-based} approaches, the importance of parameters for the previous task is measured to restrict their changes \citep{RWalk, ewc, MAS}. \textit{Regularization-based} approaches \citep{CPR, LwF} introduce regularization terms to prevent forgetting. \textit{Memory-based} approaches \citep{a-gem, cl_hindsight} keep a small episodic memory from the previous tasks and replay it to maintain knowledge. Our work is more related to the regularization-based approaches, introducing the additional local objective term to prevent forgetting out-local knowledge.

It would be worth to mention that there are some works that tried to conduct classical continual learning problems in federated learning setups. For example, \citep{FedWeIT} studied the scenario in which each local client has to learn a sequence of tasks. Here, the task-specific parameters are decomposed from the global parameters to minimize the interference between tasks. In \citep{fed_class_incremental}, the relation knowledge for the old classes is transferred round by round with class-aware gradient compensations.

\section{Server Prediction Consistency}
We extend the motivational experiment in Section 3.1 to the main experimental setups. Here, we plotted the \textit{normalized} class-wise test accuracy for each case to identify the contribution of each class on the current accuracy. This helps observe the prediction consistency regardless of the highly fluctuating global server accuracy in the non-IID case. As in \autoref{fig:appendix_server_consistency}, FedNTD effectively preserves the knowledge from the previous rounds; thereby the global server model becomes to predict each class evenly much earlier than FedAvg. Note that we normalize the class-wise test accuracy as round-by-round manner, which makes the sum for each round as 1.0.

\begin{figure}[ht!]
    \centering
    \includegraphics[width=0.95\textwidth]{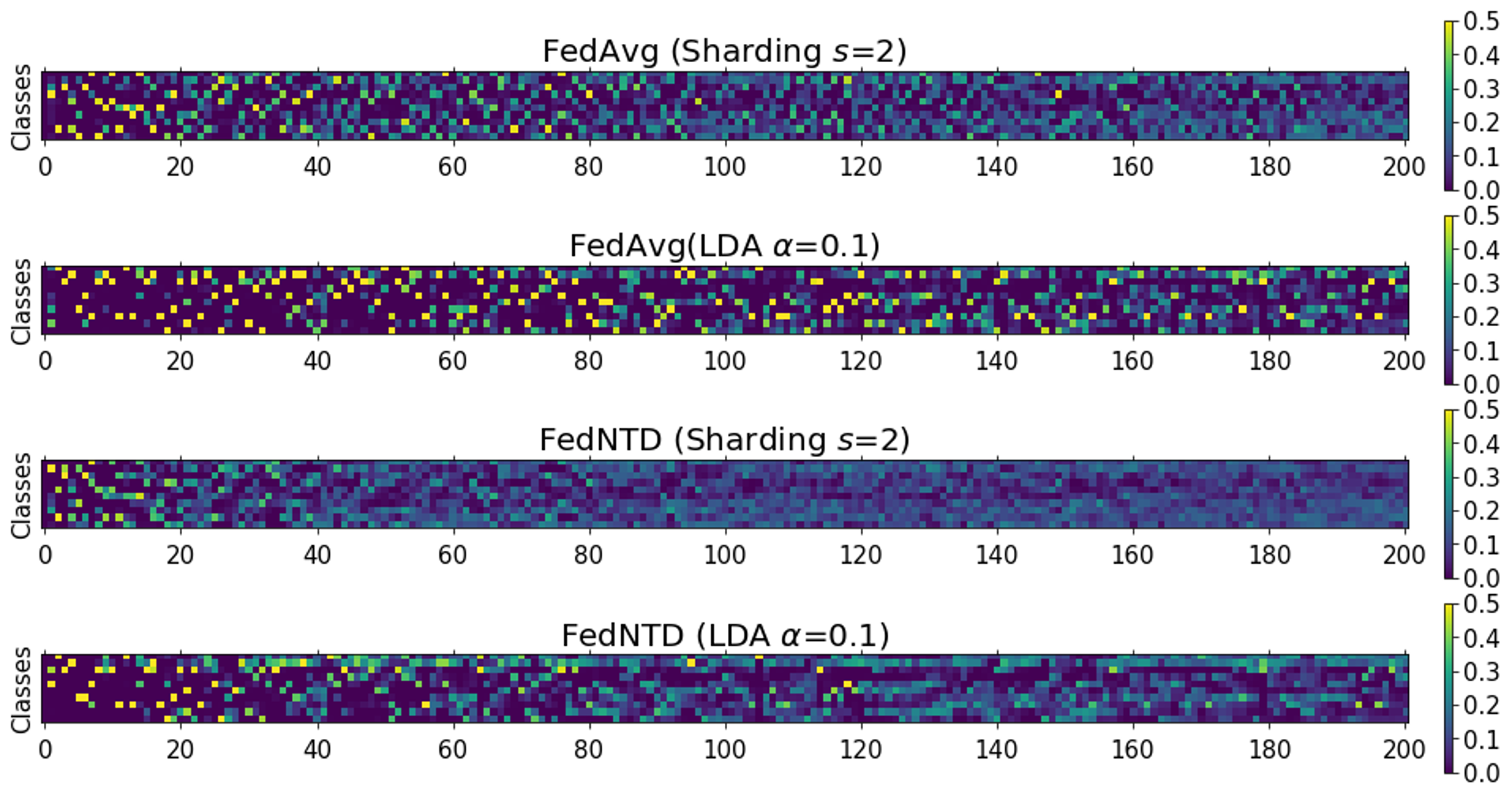}
    \caption{Visualized server test accuracy of FedAvg and FedNTD on CIFAR-10 datasets.}
    \label{fig:appendix_server_consistency}
\end{figure}
\clearpage

\section{Experiment Table with Standard Deviation}
\vspace{-10pt}
\label{appendix_table_std}
\begingroup
\setlength{\tabcolsep}{1.2pt} 
\renewcommand{\arraystretch}{1.07}
\begin{table*}[ht!]
\caption{\small Accuracy@1 (\%) on MNIST \citep{mnist}, CIFAR-10 \citep{cifar}, CIFAR-100 \citep{cifar}, and CINIC-10 \citep{cinic10}. The values in the parenthesis are the standard deviation. The arrow (\scriptsize{\textcolor{red}{$\downarrow$}},\; \scriptsize{\textcolor{green!50!black}{$\uparrow$}})\; \normalsize shows the comparison to the FedAvg.}
\small
\centering
\begin{tabular}{l|ccccccc} 
\toprule
\multicolumn{8}{c}{\textbf{NIID Partition Strategy : Sharding}} \\ 
\hline
\multicolumn{1}{c}{\multirow{2}{*}{\textbf{Method}}} & \multirow{2}{*}{\textbf{MNIST}} & \multicolumn{4}{c}{\textbf{CIFAR-10}}             
& \multirow{2}{*}{\textbf{CIFAR-100}} & \multirow{2}{*}{\textbf{CINIC-10}}  \\
\multicolumn{1}{c}{}    &       & $s=2$     & $s=3$     & $s=5$     & $s=10$    &       &  \\ 
\hline\hline
FedAvg  \scriptsize{\cite{FedAvg}} 
& 78.63${}_{\pm 0.42}\;\;$ 
& 40.14${}_{\pm 1.15}\;\;$ 
& 51.10${}_{\pm 0.11}\;\;$ 
& 57.17${}_{\pm 0.12}\;\;$  
& 64.91${}_{\pm 0.69}\;\;$ 
& 25.57${}_{\pm 0.44}\;\;$ 
& 39.64${}_{\pm 0.78}\;\;$ \\
\hline

FedCurv  \scriptsize{\cite{FedCurv}} 
& 78.56${}_{\pm 0.23}$ \scriptsize{\textcolor{red}{$\downarrow$}}  
& 44.52${}_{\pm 0.44}$ \scriptsize{\textcolor{green!50!black}{$\uparrow$}}
& 49.00${}_{\pm 0.41}$ \scriptsize{\textcolor{red}{$\downarrow$}}  
& 54.61${}_{\pm 0.20}$ \scriptsize{\textcolor{red}{$\downarrow$}}  
& 62.19${}_{\pm 0.47}$ \scriptsize{\textcolor{red}{$\downarrow$}}  
& 22.89${}_{\pm 0.66}$ \scriptsize{\textcolor{red}{$\downarrow$}}  
& 40.45${}_{\pm 0.25}$ \scriptsize{\textcolor{green!50!black}{$\uparrow$}} \\


FedProx  \scriptsize{\cite{FedProx}} 
& 78.26${}_{\pm 0.28}$ \scriptsize{\textcolor{red}{$\downarrow$}}  
& 41.48${}_{\pm 1.08}$ \scriptsize{\textcolor{green!50!black}{$\uparrow$}}
& 51.65${}_{\pm 0.53}$ \scriptsize{\textcolor{green!50!black}{$\uparrow$}}
& 56.88${}_{\pm 0.15}$ \scriptsize{\textcolor{red}{$\downarrow$}} 
& 64.65${}_{\pm 0.61}$ \scriptsize{\textcolor{red}{$\downarrow$}} 
& 25.10${}_{\pm 0.67}$ \scriptsize{\textcolor{red}{$\downarrow$}}
& 41.47${}_{\pm 0.99}$ \scriptsize{\textcolor{green!50!black}{$\uparrow$}} \\


FedNova  \scriptsize{\cite{FedNova}}  
& 77.04${}_{\pm 0.98}$ \scriptsize{\textcolor{red}{$\downarrow$}} 
& 42.62${}_{\pm 1.32}$ \scriptsize{\textcolor{green!50!black}{$\uparrow$}}
& 52.03${}_{\pm 1.49}$ \scriptsize{\textcolor{green!50!black}{$\uparrow$}}
& 62.14${}_{\pm 0.74}$ \scriptsize{\textcolor{green!50!black}{$\uparrow$}}
& 66.97${}_{\pm 0.39}$ \scriptsize{\textcolor{green!50!black}{$\uparrow$}}
& 26.96${}_{\pm 0.59}$ \scriptsize{\textcolor{green!50!black}{$\uparrow$}}
& 42.55${}_{\pm 0.10}$ \scriptsize{\textcolor{green!50!black}{$\uparrow$}} \\

SCAFFOLD \scriptsize{\cite{SCAFFOLD}}  
& 81.05${}_{\pm 0.26}$ \scriptsize{\textcolor{green!50!black}{$\uparrow$}}
& 44.60${}_{\pm 2.24}$ \scriptsize{\textcolor{green!50!black}{$\uparrow$}}
& 54.26${}_{\pm 0.22}$ \scriptsize{\textcolor{green!50!black}{$\uparrow$}} 
& \textbf{65.74${}_{\pm 0.36}$} \scriptsize{\textcolor{green!50!black}{$\uparrow$}} 
& \textbf{68.97${}_{\pm 0.34}$}  \scriptsize{\textcolor{green!50!black}{$\uparrow$}}
& 30.82${}_{\pm 0.31}$ \scriptsize{\textcolor{green!50!black}{$\uparrow$}}
& 42.66${}_{\pm 0.92}$ \scriptsize{\textcolor{green!50!black}{$\uparrow$}} \\


MOON  \scriptsize{\cite{MOON}} 
& 76.56${}_{\pm 0.24}$ \scriptsize{\textcolor{red}{$\downarrow$}}  
& 38.51${}_{\pm 0.96}$ \scriptsize{\textcolor{red}{$\downarrow$}}  
& 50.47${}_{\pm 0.15}$ \scriptsize{\textcolor{red}{$\downarrow$}}  
& 56.69${}_{\pm 0.11}$ \scriptsize{\textcolor{red}{$\downarrow$}}  
& 65.30${}_{\pm 0.51}$ \scriptsize{\textcolor{green!50!black}{$\uparrow$}}
& 25.29${}_{\pm 0.24}$ \scriptsize{\textcolor{red}{$\downarrow$}}  
& {37.07}${}_{\pm 0.24}$ \scriptsize{\textcolor{red}{$\downarrow$}}   \\

\hline
\textbf{FedNTD (Ours)}
& \textbf{84.44${}_{\pm 0.43}$} \scriptsize{\textcolor{green!50!black}{$\uparrow$}}    
& \textbf{52.61${}_{\pm 1.00}$} \scriptsize{\textcolor{green!50!black}{$\uparrow$}}
& \textbf{58.18${}_{\pm 1.42}$} \scriptsize{\textcolor{green!50!black}{$\uparrow$}}    
& 64.93${}_{\pm 0.34}$ \scriptsize{\textcolor{green!50!black}{$\uparrow$}} 
& 68.56${}_{\pm 0.24}$ \scriptsize{\textcolor{green!50!black}{$\uparrow$}} 
& \textbf{31.69${}_{\pm 0.13}$} \scriptsize{\textcolor{green!50!black}{$\uparrow$}}
& \textbf{48.07${}_{\pm 0.36}$} \scriptsize{\textcolor{green!50!black}{$\uparrow$}}\\
\hline\hline

\multicolumn{8}{c}{\textbf{NIID Partition Strategy : LDA}}  \\ 
\hline
\multicolumn{1}{c}{\multirow{2}{*}{\textbf{Method}}}  & \multirow{2}{*}{\textbf{MNIST}}  & \multicolumn{4}{c}{\textbf{CIFAR-10}}  
& \multirow{2}{*}{\textbf{CIFAR-100}}  & \multirow{2}{*}{\textbf{CINIC-10}}   \\
\multicolumn{1}{c}{} &   & $\alpha=0.05$ & $\alpha=0.1$ & $\alpha=0.3$  & $\alpha=0.5$ & &  \\ 
\hline\hline
FedAvg  \scriptsize{\cite{FedAvg}} 
& 79.73${}_{\pm 0.20}\;\;$  
& 28.24${}_{\pm 3.11}\;\;$ 
& 46.49${}_{\pm 0.93}\;\;$ 
& 57.24${}_{\pm 0.21}\;\;$  
& 62.53${}_{\pm 0.41}\;\;$  
& 30.69${}_{\pm 0.27}\;\;$ 
& 38.14${}_{\pm 3.40}\;\;$ \\
\hline

FedCurv  \scriptsize{\cite{FedCurv}} 
& 78.72${}_{\pm 0.44}$ \scriptsize{\textcolor{red}{$\downarrow$}}  
& 33.64${}_{\pm 2.98}$ \scriptsize{\textcolor{green!50!black}{$\uparrow$}}
& 44.26${}_{\pm 0.79}$ \scriptsize{\textcolor{red}{$\downarrow$}}  
& 54.93${}_{\pm 0.46}$ \scriptsize{\textcolor{red}{$\downarrow$}}  
& 59.37${}_{\pm 0.24}$ \scriptsize{\textcolor{red}{$\downarrow$}}  
& 29.16${}_{\pm 0.22}$ \scriptsize{\textcolor{red}{$\downarrow$}}  
& 36.69${}_{\pm 3.03}$ \scriptsize{\textcolor{red}{$\downarrow$}}  \\


FedProx  \scriptsize{\cite{FedProx}} 
& 79.25${}_{\pm 0.16}$ \scriptsize{\textcolor{red}{$\downarrow$}}  
& 37.19${}_{\pm 3.17}$ \scriptsize{\textcolor{green!50!black}{$\uparrow$}}
& 47.65${}_{\pm 0.90}$ \scriptsize{\textcolor{green!50!black}{$\uparrow$}} 
& 57.35${}_{\pm 0.40}$ \scriptsize{\textcolor{green!50!black}{$\uparrow$}}
& 62.39${}_{\pm 0.31}$ \scriptsize{\textcolor{red}{$\downarrow$}}  
& 30.60${}_{\pm 0.16}$ \scriptsize{\textcolor{red}{$\downarrow$}}
& 39.47${}_{\pm 3.40}$ \scriptsize{\textcolor{green!50!black}{$\uparrow$}} \\


FedNova  \scriptsize{\cite{FedNova}}  
& 60.37${}_{\pm 2.71}$ \scriptsize{\textcolor{red}{$\downarrow$}}  
& 10.00 \scriptsize{(\textit{Failed})} \scriptsize{\textcolor{red}{$\downarrow$}}
& 28.06${}_{\pm 0.12}$ \scriptsize{\textcolor{red}{$\downarrow$}} 
& 57.44${}_{\pm 1.69}$ \scriptsize{\textcolor{green!50!black}{$\uparrow$}}          
& 64.65${}_{\pm 0.34}$ \scriptsize{\textcolor{green!50!black}{$\uparrow$}} 
& 32.15${}_{\pm 0.13}$ \scriptsize{\textcolor{green!50!black}{$\uparrow$}}
& 30.44${}_{\pm 1.35}$ \scriptsize{\textcolor{red}{$\downarrow$}} \\

SCAFFOLD \scriptsize{\cite{SCAFFOLD}}  
& 71.57${}_{\pm 0.72}$ \scriptsize{\textcolor{red}{$\downarrow$}} 
& 10.00 \scriptsize{(\textit{Failed})} \scriptsize{\textcolor{red}{$\downarrow$}}
& 23.12${}_{\pm 0.55}$ \scriptsize{\textcolor{red}{$\downarrow$}} 
& \textbf{62.01${}_{\pm 0.34}$}  \scriptsize{\textcolor{green!50!black}{$\uparrow$}}
& \textbf{66.16${}_{\pm 0.13}$}  \scriptsize{\textcolor{green!50!black}{$\uparrow$}}
& 33.68${}_{\pm 0.13}$ \scriptsize{\textcolor{green!50!black}{$\uparrow$}}
& 28.78${}_{\pm 1.26}$ \scriptsize{\textcolor{red}{$\downarrow$}} \\


MOON  \scriptsize{\cite{MOON}} 
& 78.95${}_{\pm 0.46}$ \scriptsize{\textcolor{red}{$\downarrow$}}  
& 28.35${}_{\pm 3.68}$ \scriptsize{\textcolor{green!50!black}{$\uparrow$}}
& 44.77${}_{\pm 1.12}$ \scriptsize{\textcolor{red}{$\downarrow$}}  
& 58.38${}_{\pm 0.09}$ \scriptsize{\textcolor{green!50!black}{$\uparrow$}}
& 63.10${}_{\pm 0.00}$ \scriptsize{\textcolor{green!50!black}{$\uparrow$}}
& 30.64${}_{\pm 0.12}$ \scriptsize{\textcolor{red}{$\downarrow$}}  
& 37.92${}_{\pm 3.31}$ \scriptsize{\textcolor{red}{$\downarrow$}}  \\

\hline
\textbf{FedNTD (Ours)}
& \textbf{81.34${}_{\pm 0.33}$} \scriptsize{\textcolor{green!50!black}{$\uparrow$}} 
& \textbf{40.17${}_{\pm 3.19}$} \scriptsize{\textcolor{green!50!black}{$\uparrow$}}
& \textbf{54.42${}_{\pm 0.06}$} \scriptsize{\textcolor{green!50!black}{$\uparrow$}} 
& \textbf{62.42${}_{\pm 0.53}$} \scriptsize{\textcolor{green!50!black}{$\uparrow$}} 
& \textbf{66.12${}_{\pm 0.26}$} \scriptsize{\textcolor{green!50!black}{$\uparrow$}} 
& 32.37${}_{\pm 0.02}$ \scriptsize{\textcolor{green!50!black}{$\uparrow$}}
& \textbf{46.24${}_{\pm 1.67}$} \scriptsize{\textcolor{green!50!black}{$\uparrow$}}\\
\bottomrule
\end{tabular}
\label{tab:main_std}
\end{table*}
\endgroup

\section{Additional Experiments}
\vspace{-6pt}
\label{appendix_additional}

\subsection{Effect of Local Epochs}
\vspace{-10pt}

\begingroup
\setlength{\tabcolsep}{4.5pt} 
\renewcommand{\arraystretch}{1.1}
\begin{table}[ht!]
\caption{\small Accuracy@1 on CIFAR-10 (Sharding $s=2$). The value in the parenthesis is the forgetting $\mathcal{F}$.}
\small
\centering
\begin{tabular}{l|ccccc} 
\toprule
\multicolumn{6}{c}{\textbf{NIID Partition Strategy: Sharding ($s=2$)}} \\ 
\hline
\multicolumn{1}{l}{\multirow{2}{*}{\textbf{Method}}} & \multicolumn{5}{c}{\textbf{Local Epochs (E) }}                            \\
\multicolumn{1}{l}{}        & 1            & 3            & 5            & 10           & 20            \\ 
\hline\hline
FedAvg \citep{FedAvg}       
& 29.49 (0.70) \;
& 34.49 (0.64) \;
& 40.14 (0.59) \;
& 50.08 (0.49) \;
& 56.93 (0.42) \; \\
\hline
FedProx \citep{FedProx}     
& 29.44 (0.69) \scriptsize{\textcolor{red}{$\downarrow$}}
& 34.00 (0.64) \scriptsize{\textcolor{red}{$\downarrow$}}
& 41.48 (0.57) \scriptsize{\textcolor{green!50!black}{$\uparrow$}} 
& 42.74 (0.53) \scriptsize{\textcolor{red}{$\downarrow$}}
& 43.39 (0.52) \scriptsize{\textcolor{red}{$\downarrow$}} \\

FedNova \citep{FedNova}     
& 27.77 (0.71) \scriptsize{\textcolor{red}{$\downarrow$}}
& 32.00 (0.64) \scriptsize{\textcolor{red}{$\downarrow$}}
& 42.62 (0.56) \scriptsize{\textcolor{green!50!black}{$\uparrow$}} 
& 48.59 (0.50) \scriptsize{\textcolor{red}{$\downarrow$}}
& 58.24 (0.39) \scriptsize{\textcolor{green!50!black}{$\uparrow$}}  \\

SCAFFOLD \citep{SCAFFOLD}   
& 34.46 (0.64) \scriptsize{\textcolor{green!50!black}{$\uparrow$}} 
& 39.26 (0.58) \scriptsize{\textcolor{green!50!black}{$\uparrow$}} 
& 44.60 (0.53) \scriptsize{\textcolor{green!50!black}{$\uparrow$}} 
& 55.35 (0.41) \scriptsize{\textcolor{green!50!black}{$\uparrow$}} 
& \textbf{62.80} (0.34) \scriptsize{\textcolor{green!50!black}{$\uparrow$}}  \\
\hline
\textbf{FedNTD (Ours)}      
& \textbf{35.77} (0.64) \scriptsize{\textcolor{green!50!black}{$\uparrow$}} 
& \textbf{45.47} (0.50) \scriptsize{\textcolor{green!50!black}{$\uparrow$}} 
& \textbf{52.61} (0.43) \scriptsize{\textcolor{green!50!black}{$\uparrow$}} 
& \textbf{60.22} (0.36) \scriptsize{\textcolor{green!50!black}{$\uparrow$}} 
& 60.61 (0.34) \scriptsize{\textcolor{green!50!black}{$\uparrow$}}  \\
\bottomrule
\end{tabular}
\label{tab:local_epochs_sharding2}
\end{table}
\endgroup

\begingroup
\setlength{\tabcolsep}{4.5pt} 
\renewcommand{\arraystretch}{1.1}
\begin{table}[ht!]
\caption{\small Accuracy@1 on CIFAR-10 (LDA $\alpha=0.1$). The value in the parenthesis is the forgetting $\mathcal{F}$.}
\small
\centering
\begin{tabular}{l|ccccc} 
\toprule
\multicolumn{6}{c}{\textbf{NIID Partition Strategy: LDA ($\alpha=0.1$)}} \\ 
\hline
\multicolumn{1}{l}{\multirow{2}{*}{\textbf{Method}}} & \multicolumn{5}{c}{\textbf{Local Epochs (E) }}                            \\
\multicolumn{1}{l}{}        & 1            & 3            & 5            & 10           & 20            \\ 
\hline\hline
FedAvg \citep{FedAvg}       
& 29.77 (0.69) \;
& 37.70 (0.60) \;
& 46.49 (0.51) \;
& 53.80 (0.43) \;
& 57.70 (0.39) \; \\
\hline
FedProx \citep{FedProx}     
& 33.37 (0.65) \scriptsize{\textcolor{green!50!black}{$\uparrow$}} 
& 37.88 (0.57) \scriptsize{\textcolor{green!50!black}{$\uparrow$}} 
& 47.65 (0.49) \scriptsize{\textcolor{green!50!black}{$\uparrow$}} 
& 44.02 (0.50) \scriptsize{\textcolor{red}{$\downarrow$}} 
& 44.98 (0.49) \scriptsize{\textcolor{red}{$\downarrow$}}  \\

FedNova \citep{FedNova}     
& 26.35 (0.73) \scriptsize{\textcolor{red}{$\downarrow$}}
& 24.37 (0.74) \scriptsize{\textcolor{red}{$\downarrow$}}
& 28.06 (0.71) \scriptsize{\textcolor{red}{$\downarrow$}}
& 47.41 (0.50) \scriptsize{\textcolor{red}{$\downarrow$}}
& 10.00 (\small{\textit{Failure}}) \scriptsize{\textcolor{red}{$\downarrow$}} \\

SCAFFOLD \citep{SCAFFOLD}   
& 13.36 (0.86) \scriptsize{\textcolor{red}{$\downarrow$}}
& 22.04 (0.75) \scriptsize{\textcolor{red}{$\downarrow$}}
& 23.12 (0.74) \scriptsize{\textcolor{red}{$\downarrow$}}
& 38.49 (0.57)\scriptsize{\textcolor{red}{$\downarrow$}}
& 47.07 (0.47) \scriptsize{\textcolor{red}{$\downarrow$}} \\
\hline
\textbf{FedNTD (Ours)}      
& \textbf{33.94} (0.64) \scriptsize{\textcolor{green!50!black}{$\uparrow$}} 
& \textbf{45.92} (0.50) \scriptsize{\textcolor{green!50!black}{$\uparrow$}} 
& \textbf{54.42} (0.42) \scriptsize{\textcolor{green!50!black}{$\uparrow$}} 
& \textbf{60.67} (0.33) \scriptsize{\textcolor{green!50!black}{$\uparrow$}} 
& \textbf{62.25} (0.30) \scriptsize{\textcolor{green!50!black}{$\uparrow$}}  \\
\bottomrule
\end{tabular}
\label{tab:local_epochs_lda01}
\end{table}
\endgroup

\clearpage
\subsection{Effect of Sampling Ratio}
\vspace{-10pt}

\begingroup
\setlength{\tabcolsep}{4.5pt} 
\renewcommand{\arraystretch}{1.1}
\begin{table}[ht!]
\caption{\small Accuracy@1 on CIFAR-10 (Sharding $s=2$). The value in the parenthesis is the forgetting $\mathcal{F}$.}
\small
\centering
\begin{tabular}{l|ccccc} 
\toprule
\multicolumn{6}{c}{\textbf{NIID Partition Strategy: Sharding ($s=2$)}} \\ 
\hline
\multicolumn{1}{l}{\multirow{2}{*}{\textbf{Method}}} & \multicolumn{5}{c}{\textbf{Client Sampling Ratio (R)}}  \\
\multicolumn{1}{l}{}        & 0.05          & 0.1            & 0.3            & 0.5     & 1.0\\ 
\hline\hline
FedAvg \citep{FedAvg}       
& 33.06 (0.66) \;
& 40.14 (0.59) \;
& 49.99 (0.46) \;
& 52.98 (0.41) \;   
& 51.48 (0.30) \;
\\
\hline
FedProx \citep{FedProx}     
& 35.36 (0.63) \scriptsize{\textcolor{green!50!black}{$\uparrow$}} 
& 41.48 (0.57) \scriptsize{\textcolor{green!50!black}{$\uparrow$}} 
& 44.54 (0.45) \scriptsize{\textcolor{red}{$\downarrow$}}
& 50.02 (0.31) \scriptsize{\textcolor{red}{$\downarrow$}}
& 52.53 (0.06) \scriptsize{\textcolor{green!50!black}{$\uparrow$}} 
\\

FedNova \citep{FedNova}     
& 29.99 (0.69) \scriptsize{\textcolor{red}{$\downarrow$}} 
& 42.62 (0.56) \scriptsize{\textcolor{green!50!black}{$\uparrow$}} 
& 55.59 (0.31) \scriptsize{\textcolor{green!50!black}{$\uparrow$}} 
& 56.75 (0.23) \scriptsize{\textcolor{green!50!black}{$\uparrow$}}    
& 51.89 (0.34) \scriptsize{\textcolor{green!50!black}{$\uparrow$}} 
\\

SCAFFOLD \citep{SCAFFOLD}   
& 29.15 (0.70) \scriptsize{\textcolor{red}{$\downarrow$}}
& 44.60 (0.53) \scriptsize{\textcolor{green!50!black}{$\uparrow$}} 
& 55.59 (0.31) \scriptsize{\textcolor{green!50!black}{$\uparrow$}} 
& 56.75 (0.23) \scriptsize{\textcolor{green!50!black}{$\uparrow$}}  
& 57.88 (0.10) \scriptsize{\textcolor{green!50!black}{$\uparrow$}}  
\\
\hline
\textbf{FedNTD (Ours)}      
& \textbf{46.99} (0.51) \scriptsize{\textcolor{green!50!black}{$\uparrow$}}  
& \textbf{52.61} (0.43) \scriptsize{\textcolor{green!50!black}{$\uparrow$}} 
& \textbf{59.37} (0.28) \scriptsize{\textcolor{green!50!black}{$\uparrow$}} 
& \textbf{60.70} (0.18) \scriptsize{\textcolor{green!50!black}{$\uparrow$}} 
& \textbf{61.53} (0.04) \scriptsize{\textcolor{green!50!black}{$\uparrow$}} 
\\
\bottomrule
\end{tabular}
\label{tab:sampling_ratio_sharding2}
\end{table}
\endgroup

\vspace{-15pt}
\begingroup
\setlength{\tabcolsep}{4.5pt} 
\renewcommand{\arraystretch}{1.1}
\begin{table}[ht!]
\caption{\small Accuracy@1 on CIFAR-10 (LDA $\alpha=0.1$). The value in the parenthesis is the forgetting $\mathcal{F}$.}
\small
\centering
\begin{tabular}{l|ccccc} 
\toprule
\multicolumn{6}{c}{\textbf{NIID Partition Strategy: LDA ($\alpha=0.1$)}} \\ 
\hline
\multicolumn{1}{l}{\multirow{2}{*}{\textbf{Method}}} & \multicolumn{5}{c}{\textbf{Client Sampling Ratio (R)}}  \\
\multicolumn{1}{l}{}        & 0.05          & 0.1            & 0.3            & 0.5     & 1.0\\ 
\hline\hline
FedAvg \citep{FedAvg}       
& 29.35 (0.70) \;
& 46.49 (0.51) \;
& 53.73 (0.39) \;
& 58.72 (0.25) \;   
& 61.38 (0.04) \;
\\
\hline
FedProx \citep{FedProx}     
& 36.36 (0.63) \scriptsize{\textcolor{green!50!black}{$\uparrow$}} 
& 47.65 (0.49) \scriptsize{\textcolor{green!50!black}{$\uparrow$}} 
& 45.78 (0.37) \scriptsize{\textcolor{red}{$\downarrow$}}
& 49.65 (0.23) \scriptsize{\textcolor{red}{$\downarrow$}}   
& 51.31 (0.07)) \scriptsize{\textcolor{red}{$\downarrow$}}    
\\
FedNova \citep{FedNova}     
& 21.31 (0.78) \scriptsize{\textcolor{red}{$\downarrow$}}
& 28.06 (0.71) \scriptsize{\textcolor{red}{$\downarrow$}}
& 45.83 (0.49) \scriptsize{\textcolor{red}{$\downarrow$}}
& 55.09 (0.50) \scriptsize{\textcolor{red}{$\downarrow$}}  
& 56.79 (0.30) \scriptsize{\textcolor{red}{$\downarrow$}}  
\\

SCAFFOLD \citep{SCAFFOLD}   
& 15.80 (0.84)  \scriptsize{\textcolor{red}{$\downarrow$}}
& 23.12 (0.74) \scriptsize{\textcolor{red}{$\downarrow$}}
& 41.29 (0.51) \scriptsize{\textcolor{red}{$\downarrow$}}
& 10.00 (\small{\textit{Failure}})\scriptsize{\textcolor{red}{$\downarrow$}}
& 10.00 (\small{\textit{Failure}})\scriptsize{\textcolor{red}{$\downarrow$}} \\

\hline
\textbf{FedNTD (Ours)}      
& \textbf{45.80} (0.53) \scriptsize{\textcolor{green!50!black}{$\uparrow$}}  
& \textbf{54.42} (0.42) \scriptsize{\textcolor{green!50!black}{$\uparrow$}} 
& \textbf{58.57} (0.33) \scriptsize{\textcolor{green!50!black}{$\uparrow$}} 
& \textbf{60.88} (0.19) \scriptsize{\textcolor{green!50!black}{$\uparrow$}}
& \textbf{62.48} (0.06) \scriptsize{\textcolor{green!50!black}{$\uparrow$}} \\
\bottomrule
\end{tabular}
\label{tab:sampling_ratio_lda01}
\end{table}
\endgroup

\subsection{Results on ResNet-10 Model}
We report an additional experiment on popular architecture, ResNet-10. The number of parameters in ResNet-10 is about 10x larger than the 2-conv + 2-fc model for the main experiments.
\begingroup
\setlength{\tabcolsep}{3.5pt} 
\renewcommand{\arraystretch}{1.0}
\begin{table}[h!]
\footnotesize
\centering
\begin{tabular}{c|c:cc:c} 
\toprule
\multicolumn{1}{c}{}    & \multicolumn{1}{c}{\textbf{FedAvg}}  & \textbf{Scaffold} & \multicolumn{1}{c}{\textbf{MOON}} & \textbf{FedNTD (ours)}\\ 
\hline
\textbf{Shard (s = 2)}        & 36.01    & 44.59             & 35.21                             & \textbf{46.27}   \\
\textbf{Shard (s = 5)}        & 39.21    & 65.08             & 51.02                             & \textbf{65.92}   \\
\textbf{LDA ($\alpha$ = 0.1)}    & 33.35  & 38.78             & 33.57                             & \textbf{49.85}   \\
\bottomrule
\end{tabular}
\end{table}
\endgroup

\section{Comparison to KD}
\label{appendix_comparision_kd}
\vspace{-3pt}

We analyze the advantage of FedNTD over KD by observing the performance of the loss function below. Note that $L(\lambda)$ moves from $\mathcal{L}_{\text{KD}}$ to $\mathcal{L}_{\text{NTD}}$ as $\lambda$ increases, and collapses to $\mathcal{L}_{\text{KD}}$ at $\lambda=0$ and $\mathcal{L}_{\text{NTD}}$ at  and $\lambda=1$. 
\begin{equation}
    \mathcal{L}_{\scriptscriptstyle\text{KD}\rightarrow\text{NTD}} = \mathcal{L}_{\text{CE}}(q, \; \mathds{1}_y) + L(\lambda),
\end{equation}
\begin{equation}
    L(\lambda) =  (1-\lambda)\cdot\mathcal{L}_{\text{KD}}(q_{\tau}^l, \; q_{\tau}^{g}) + \lambda\cdot \mathcal{L}_{\text{NTD}}({\tilde{q}_{\tau}^l}, \; {\tilde{q}_{\tau}^{g}}).
\end{equation}

The result is in \autoref{tab:kd_ntd}, which shows reaching $L(\lambda)$ to $\mathcal{L}_{NTD}$ significantly improves the performance. This improvement supports the effect of decoupling the not-true classes and the true classes: preservation of out-local distribution knowledge using not-true class signals and acquisition of new knowledge on true classes from local datasets.

\begingroup
\setlength{\tabcolsep}{3.0pt} 
\renewcommand{\arraystretch}{1.1}
\begin{table}[ht!]
\caption{\small CIFAR-10 test accuracy by varying $\lambda$.}
\footnotesize
\centering
\begin{tabular}{c|ccccccc} 
\toprule
\multicolumn{1}{c}{\multirow{2}{*}{\textbf{Partition Method}}} 
& \multirow{2}{*}{\textbf{KD}} 
& \multicolumn{5}{c}{\textbf{KD $\rightarrow$ NTD}} 
& \multirow{2}{*}{\textbf{NTD}}  \\
\multicolumn{1}{c}{}&       & 0.1   & 0.3   & 0.5   & 0.7   & 0.9   &       \\ 
\hline\hline
Sharding \footnotesize{($s=2$)} \normalsize    & 46.2 & 46.4 & 46.9 & 47.6 & 48.8 & 50.2 & \textbf{52.6} \\
LDA \footnotesize{($\alpha=0.1$)} \normalsize  & 50.8 & 50.9 & 51.4 & 51.9 & 52.6 & 53.6 & \textbf{54.9} \\
\bottomrule
\end{tabular}
\label{tab:kd_ntd}
\end{table}
\endgroup

\clearpage

 To further analyze the effect of NTD, we measure the performance of the local model as the communication rounds proceed. The result is plotted in \autoref{fig:kd_ntd_avg}. Note that the \textit{Personalized} performance is evaluated on the test samples with the same label distribution of the local clients.

The result shows that although the KD considerably improves the server model performance, the local model learned by KD shows much lower local performance. On the other hand, FedNTD shows much higher local performance, which implies that NTD successfully tackles the distillation not to hinder the local learning.

We insist that such significant improvement by discarding true-class logits in the distillation loss comes from the better trade-off between attaining new knowledge from local data and preserving old knowledge in the global model, as suggested in Section 3.

\begin{figure}[ht!]
    \centering
    \includegraphics[width=1.0\textwidth]{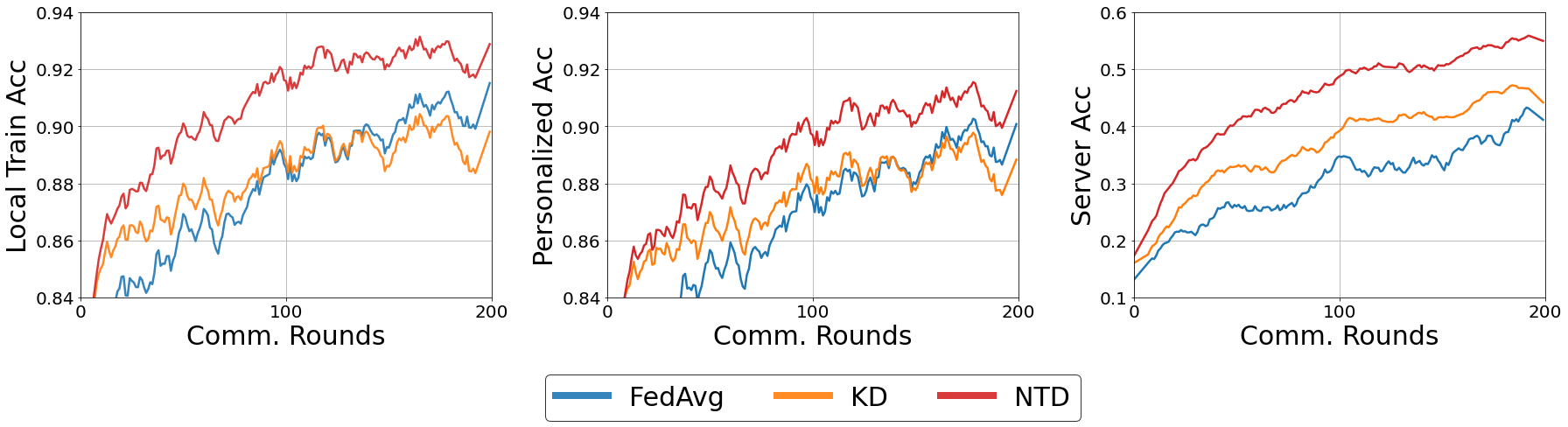}
\caption{CIFAR-10 (Sharding: $s=2$) performances from KD and NTD}
\label{fig:kd_ntd_avg}
\end{figure}

\vspace{-5pt}
\section{Personalized performance of FL methods}
\label{appendix_personalized}
\vspace{-5pt}

Here we investigate the personalized performance of our FedNTD. As suggested in \autoref{appendix_comparision_kd}, although FedNTD aims to improve global convergence, it also improves personalized performance. However, as the learning curves of SCAFFOLD (\textit{the green line}) show, the lower local learning performance does not always lead to the worse server model performance. In all cases, SCAFFOLD shows significantly lower local performance at each round (\textit{the 1st and 2nd row}), but it considerably improves the global convergence (\textit{the 3rd row}).

\begin{figure}[ht!]
    \centering
    \includegraphics[width=1.0\textwidth]{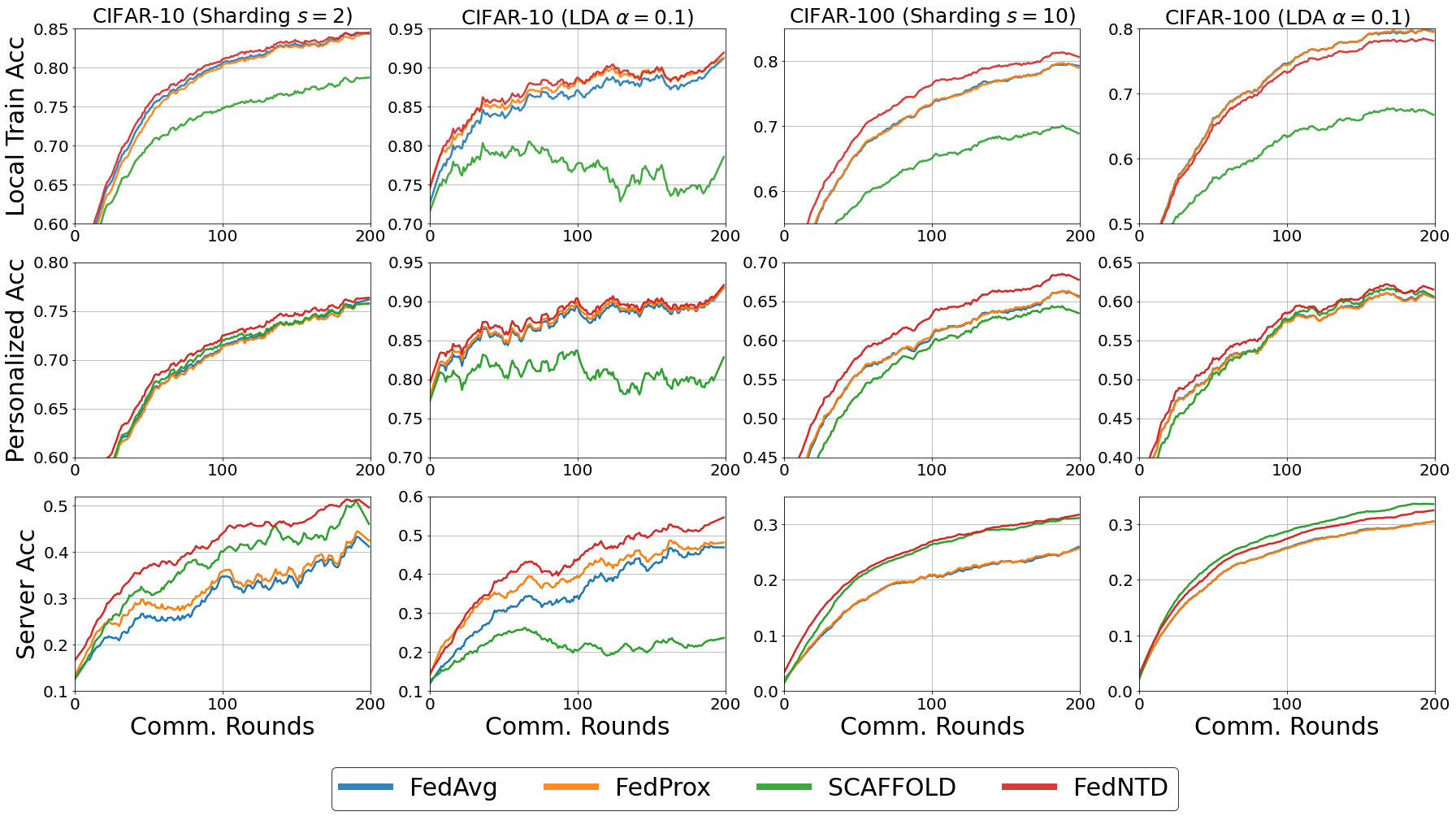}
\caption{Local and global learning curves of FL methods. The accuracy of the local model is evaluated on: \textit{(Local Train)} - the local private trains samples, and \textit{(Personalized)}: the test samples from the same label distribution with the local client}

\label{fig:personalized}
\end{figure}

\clearpage
\section{Comparison to FedAlign}

Here we compare our FedNTD with a recently proposed method, FedAlign \citep{FedAlign}, which shares the motivation of our work that local learning is the bottleneck of FL performance. In FedAlign, a correction term is introduced in the local learning target to obtain local models that generalize well. We implemented our FedNTD on officially released FedAlign code \footnote{https://github.com/mmendiet/FedAlign}, and used the hyperparameters specified in \citep{FedAlign}. The results are in \autoref{tab:fedalign} and \autoref{fig:fedalign_experiment} shows their corresponding learning curves. In our experiment, although FedAlign improves the performance at some settings (LDA $\alpha=0.5$), its learning suffers when the heterogeneity level becomes severe. On the other hand, FedNTD consistently improves the performance even in such cases.

\begin{figure}[ht!]
    \centering
    \includegraphics[width=1.0\textwidth]{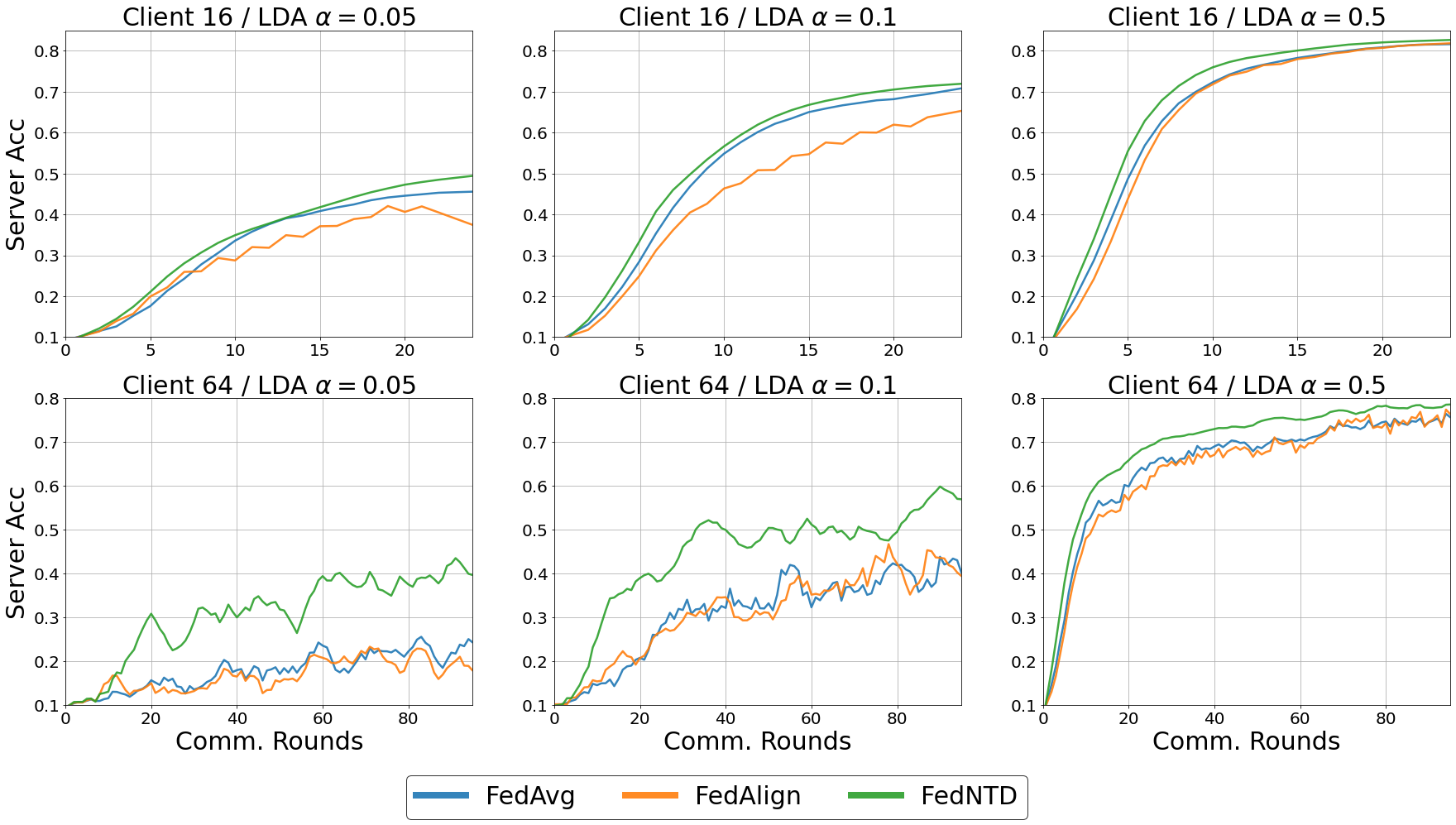}
\caption{Learning curves that corresponds to \autoref{tab:fedalign}.}
\label{fig:fedalign_experiment}
\end{figure}

\begingroup
\setlength{\tabcolsep}{6.0pt} 
\renewcommand{\arraystretch}{1.15}
\begin{table}[ht!]
\caption{CIFAR-10 test accuracy. 16 clients participates for each communication round. The number local epochs is 20 for all experiments.}
\centering
\begin{tabular}{llll}
\toprule
\textbf{Client Number (LDA $\alpha$)} & \textbf{FedAvg} \citep{FedAvg} & \textbf{FedAlign}\citep{FedAlign} & \textbf{FedNTD (ours)}  \\ 
\hline\hline
Client 16 ($\alpha=0.05$)             & 0.4556          & 0.3743            & \textbf{0.4943}  \\
Client 16 ($\alpha=0.1$)            & 0.7083          & 0.6532            & \textbf{0.7195}  \\
Client 16 ($\alpha=0.5$)            & 0.8163          & 0.8185            & \textbf{0.8266}  \\
\hline
Client 64 ($\alpha=0.05$)           & 0.2535          & 0.1854            & \textbf{0.3927}  \\
Client 64 ($\alpha=0.1$)            & 0.4247          & 0.3931            & \textbf{0.5634}  \\
Client 64 ($\alpha=0.5$)            & 0.7568          & 0.7698            & \textbf{0.7846}  \\
\bottomrule
\end{tabular}
\label{tab:fedalign}
\end{table}
\endgroup
\vspace{10pt}

 The introduced loss term of FedAlign aims to seek out-of-distribution generality w.r.t. global distribution during local training, resulting in the smooth loss landscape across domains (= heterogeneous local distributions in FL context). In \autoref{fig:loss_space}, we analyze the loss landscape using the parameter perturbation with Gaussian noise and the visualization using top-2 eigenvector axis, as in \citep{FedAlign}. Interestingly, our FedNTD also smoothed the local landscape, implying that the local training does not require significant parameter change to fit its local distribution. We expect that one can get insight into the intriguing property in the loss space geometry to tackle the data heterogeneity problem for future work.

\clearpage
\begin{figure*}[ht!]
    \centering
    \hfill
    \begin{subfigure}[b]{0.3\textwidth}  
        \centering 
        \includegraphics[width=\textwidth]{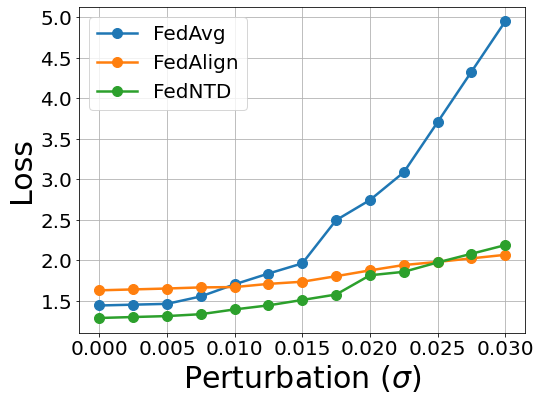}
        \caption[]%
        {{Loss change by perturbation}}
    \end{subfigure}
    \hfill
    \begin{subfigure}[b]{0.65\textwidth}  
        \centering 
        \includegraphics[width=\textwidth]{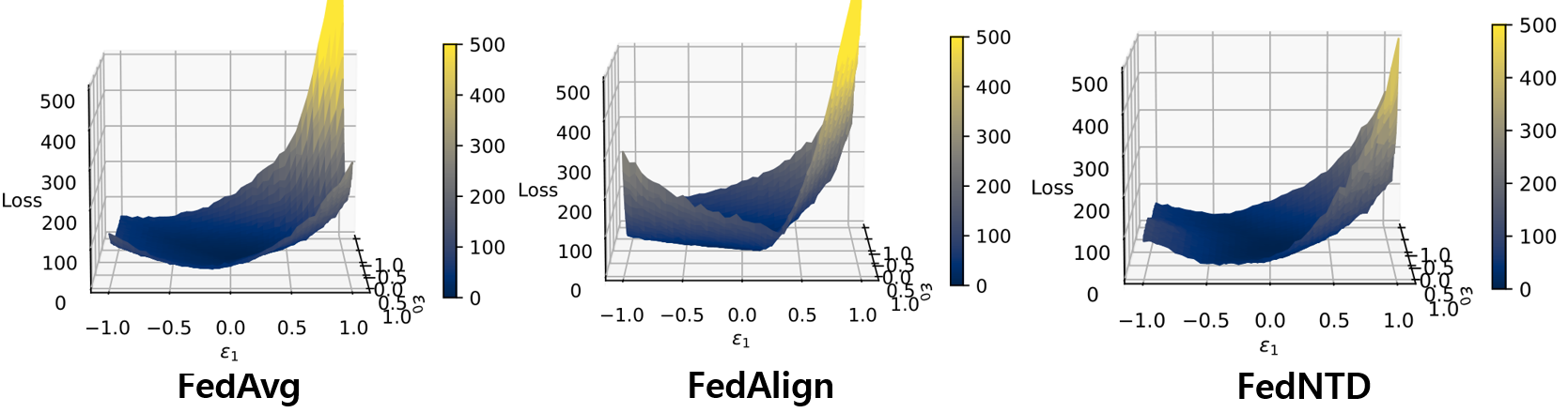}
        \caption[]%
        {{Visualized loss landscape with Hessian eigenvectors $\epsilon_0$ and $\epsilon_1$.}}
    \end{subfigure}
    \caption{Loss space of learned model (Client 16 / LDA $\alpha=0.5$).}
    \label{fig:loss_space}
\end{figure*}

\vspace{-10pt}
\section{Effect of FedNTD Hyperparameters}
\vspace{-5pt}
\label{appendix_ntd_tau_beta}

In \autoref{fig:tau_beta}, we plot the effect of FedNTD hyperparameters on the performance. The result shows that although FedNTD is not much sensitive to the choice of $\beta$, too small $\tau$ significantly drops the accuracy., which may be due to the too stiff not-true probability targets. The effect of both hyperparameters on the forgetting measure $\mathcal{F}$ is in \autoref{fig:f_tau_beta}.
\vspace{-5pt}

\begin{figure}[ht!]
    \centering
    \includegraphics[width=0.7\textwidth]{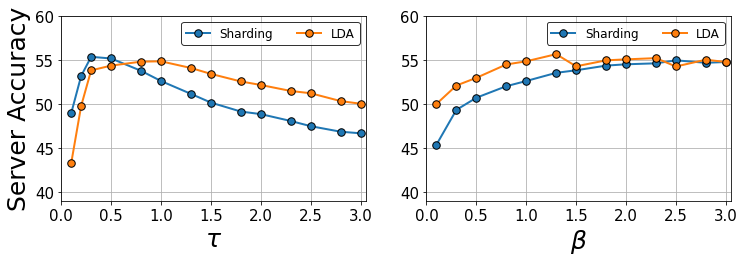}
    \vspace{-6pt}
\caption{CIFAR-10 (Sharding: $s=2$, LDA: $\alpha=0.1$) test accuracy by varying FedNTD hyperparameter $\tau$ and $\beta$ values.}
\label{fig:tau_beta}
\end{figure}
\vspace{-8pt}

\begin{figure}[ht!]  
    \centering
    \includegraphics[width=0.75\textwidth]{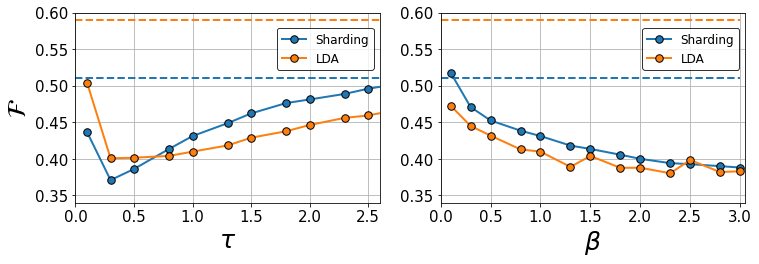}
    \vspace{-6pt}
\caption{Forgetting $\mathcal{F}$ of FedNTD on CIFAR-10 by varying hyperparameters. The dotted lines stands for the baseline FedAvg.}
\label{fig:f_tau_beta}
\end{figure}
\vspace{-10pt}

\section{MSE Loss for Not-True Distillation}
\vspace{-5pt}
\label{appendix_prop2_mse}
We explore how the MSE loss on Not-True Distillation acts. In \autoref{tab:mse}, the MSE version FedNTD (FedNTD (MSE)) shows better accuracy and less forgets as $\beta$ grows, but at some degree, the model diverges; thereby cannot reach the original FedNTD, which exploits softmax and KL-Divergence loss to distill the knowledge in the global model. We explain it as matching all not-true logits using MSE logits is too strict to learn the global knowledge since the dark knowledge is mainly contained in top-k logits. FedNTD controls the class signals by using temperature-softened softmax.

\vspace{-10pt}
\begingroup
\setlength{\tabcolsep}{3.0pt} 
\renewcommand{\arraystretch}{1.1}
\begin{table}[ht!]
\caption{CIFAR-10 (Sharding s=2) results by varying $\beta$ for FedNTD (MSE) and FedNTD}
\label{tab:mse}
\centering
\begin{tabular}{ccccccccc} 
\toprule
\textbf{Method}       & \textbf{FedAvg} & \multicolumn{6}{c}{\textbf{FedNTD (MSE)}}       & \textbf{FedNTD}  \\ 
\hline\hline
\textbf{$\beta$}         & 0.0               & 0.001 & 0.005 & 0.01  & 0.05  & 0.1   & 0.3     & 1.0                \\ 
\hline
\textbf{Accuracy}     & 40.14           & 40.53 & 42.39 & 43.02 & 44.41 & 44.27 & \textit{Failure} & \underline{\textbf{52.61}}   \\
\textbf{Forgetting $\mathcal{F}$} & 0.59            & 0.58  & 0.56  & 0.55  & 0.53  & 0.53  & \textit{Failure}       & \underline{\textbf{0.43}}    \\
\bottomrule
\end{tabular}
\end{table}
\endgroup

\clearpage
\section{Visualization of Feature Alignment}
\vspace{-3pt}

\label{appendix_feature_alignment}
To analyze feature alignment, we regard a neuron as the basic feature unit and identify individual neuron's class preference as follows:

\vspace{-18pt}
\begin{equation}
    \mathcal{H} = [h_1, h_2 \dots {h}_{\mathcal{C}}], \:\: \text{where}\:\:{h}_{c} = \sum_{i=1}^{N_c} \mathcal{O}(x_{c,i}).
\end{equation}
\vspace{-15pt}

Here, the $\mathcal{O}(x_{c,i})$ denotes the neuron's activation on data $x_i$ of class $c$, and $N_c$ is the number of samples for class $c$. For each neuron, we obtain the largest class index $\text{argmax}_i(\mathcal{H}_i)$, to identify the most dominantly encoded class semantics. A similar measure is adopted in \cite{Fed2}. In \autoref{fig:appendix_feature_alignment}, we visualize the last layer neurons' class preference. In both IID and NIID (Sharding $s=2$) cases, the features are more well-aligned in FedNTD.

\begin{figure}[ht!]
    \centering
    \includegraphics[width=0.8\textwidth]{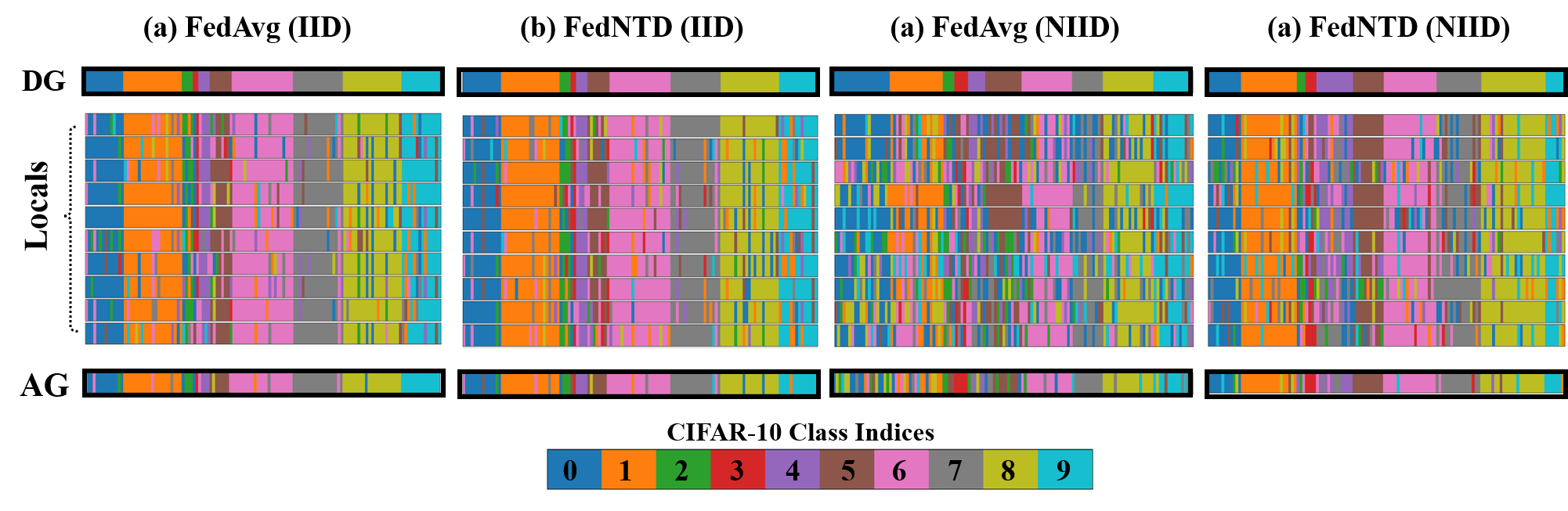}
    \caption{Visualized server test accuracy of FedAvg and FedNTD on CIFAR-10 (Sharding=2).}
    \label{fig:appendix_feature_alignment}
\end{figure}

\section{Local features visualization: Hypersphere}
\vspace{-3pt}
\label{appendix_hypersphere}
To figure out forgetting of knowledge in the global model, we now analyze how the representation on the global distribution changes during local training. To this end, we design a straightforward experiment that shows the change of features on the unit hypersphere. More specifically, we modified the network architecture to map CIFAR-10 (Sharding $s=2$) input data to 2-dimensional vectors and normalize them to be aligned on the unit hypersphere $S^{1} = \{x \in \mathbb{R}^2: {||x||}_2 = 1\}$. We then estimate their probability density function. The global model is learned for 100 rounds of communication on \textit{homogeneous} locals (i.i.d. distributed) and distributed to \textit{heterogeneous} locals with different local distributions. The result is in \autoref{fig:feature_ring}

\begin{figure}[ht!]
    \centering
    \includegraphics[width=0.9\textwidth]{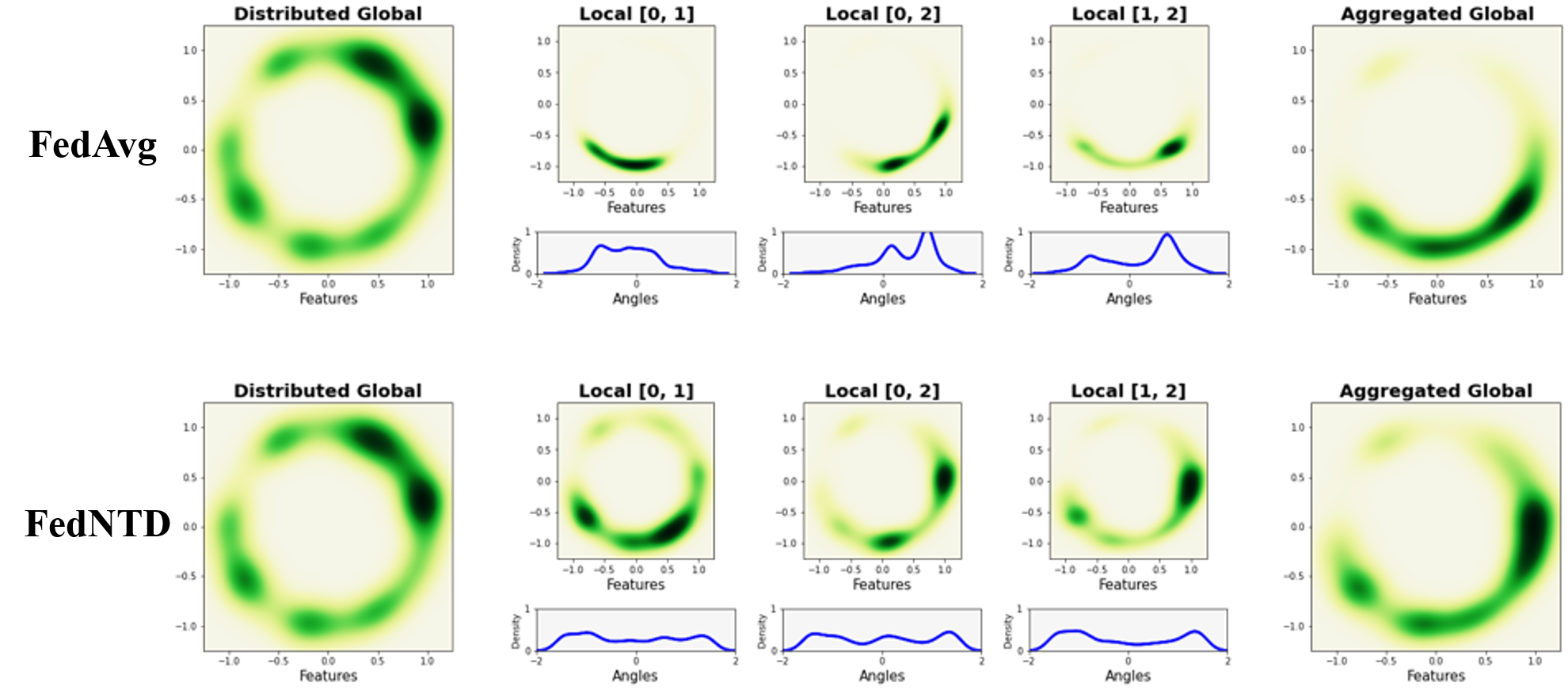}
    \vspace{-5pt}
    \caption{Features of CIFAR-10 (Sharding $s=2$) test samples on $S^2$. We plot the feature distribution with Gaussian kernel density estimation (KDE) in $\mathbb{R}^2$ and $\arctan(y, x)$ for each point $(x, y) \in S^1$. The distributed global model (first column) is trained on heterogeneous locals (middle 3 columns) and aggregated by parameter averaging (last column).}
    \label{fig:feature_ring}
\end{figure}

\clearpage
\section{Local features visualization: T-SNE}
\label{appendix_tsne}
\vspace{-6pt}
We further conduct an additional experiment on features of the trained local model. we trained the global server model for 100 communication rounds on \textit{heterogeneous} (NIID) locals and distributed over 10 \textit{homogeneous} (IID) locals and 10 \textit{heterogeneous} (NIID) locals. In the homogeneous local case (\autoref{fig:tsne_homogeneous_class}, \autoref{fig:tsne_homogeneous_local}), the features are clustered by classes, regardless of which local they are learned from. On the other hand, in the heterogeneous local case (\autoref{fig:tsne_heterogeneous_class}, \autoref{fig:tsne_heterogeneous_local}), the features are clustered by which local distribution is learned. In \autoref{fig:tsne_avg_ntd}, we visualize the effect of FedNTD on the local features.
\vspace{-3pt}
\begin{figure*}[ht!]
    \centering
    \hfill
    \begin{subfigure}[b]{0.43\textwidth}  
        \centering
        \includegraphics[width=\textwidth]{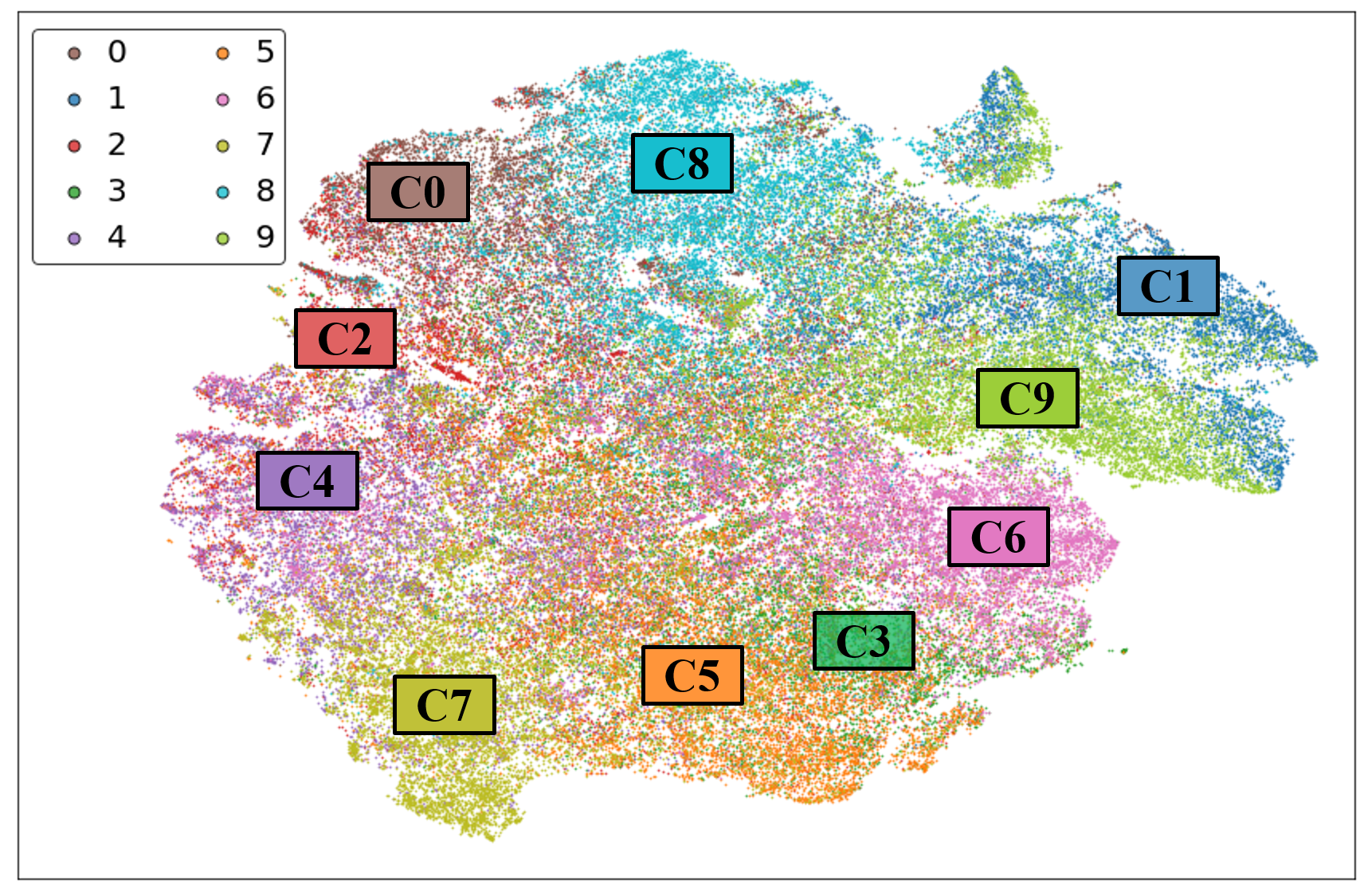}
        \caption[]%
        {{\small Homogeneous Locals (colored by \underline{\textit{classes}}})}
        \label{fig:tsne_homogeneous_class}
    \end{subfigure}
    \hfill
    \begin{subfigure}[b]{0.43\textwidth}  
        \centering 
        \includegraphics[width=\textwidth]{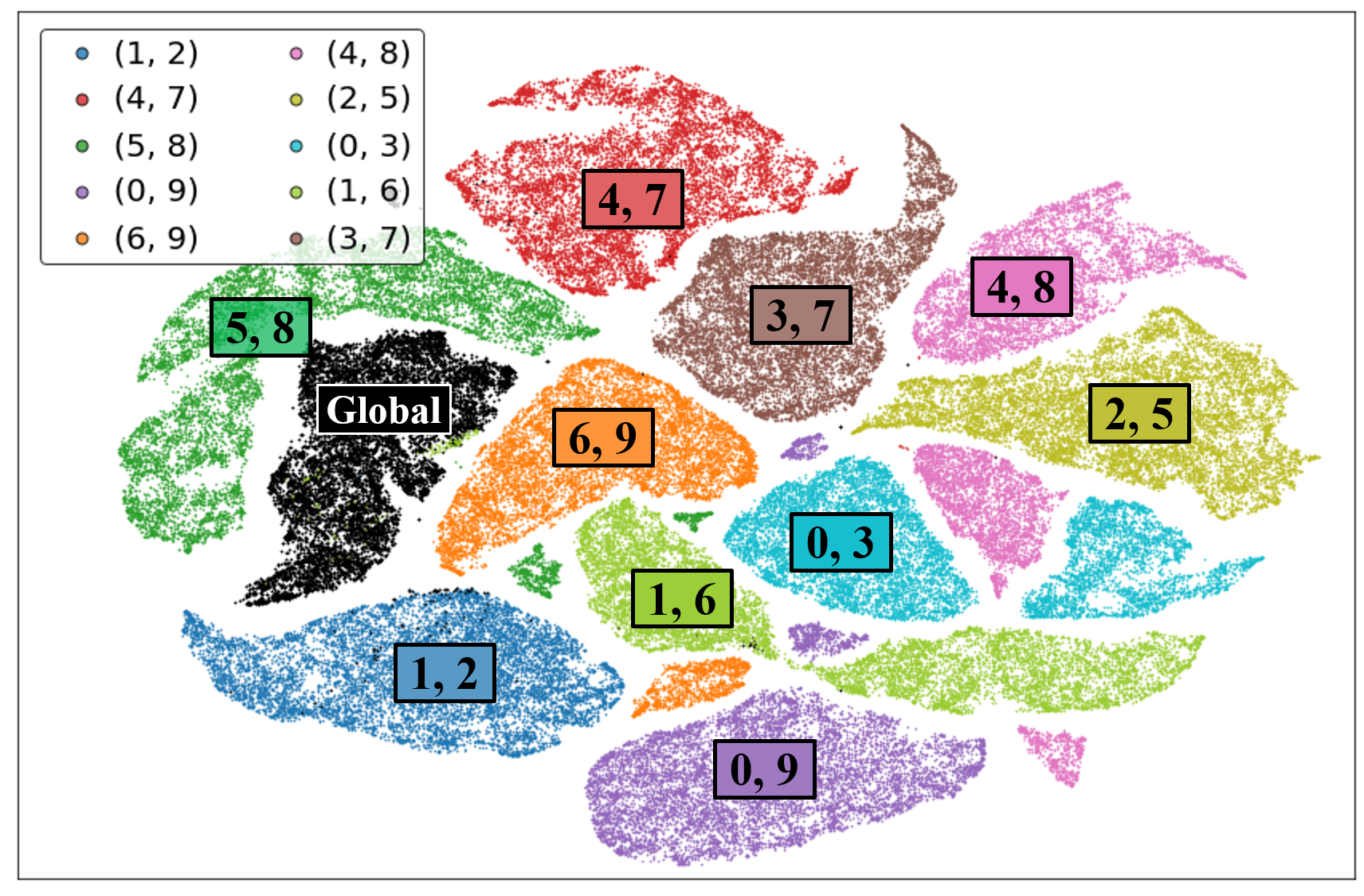}
        \caption[]%
        {{\small Heterogeneous Locals (colored by \underline{\textit{locals}}})}
        \label{fig:tsne_heterogeneous_class}
    \end{subfigure}
    \caption{\small T-SNE visualization of features on CIFAR-10 test samples after local training on (a) \textit{homogeneous} local distributions and (b) \textit{heterogeneous} local distributions. The T-SNE is conducted together for the test sample features of global and 10 local models.}
    \label{fig:local_vis1}
\end{figure*}
\vspace{-8pt}
\begin{figure*}[ht!]
    \centering
    \begin{subfigure}[b]{0.43\textwidth}  
        \centering
        \includegraphics[width=\textwidth]{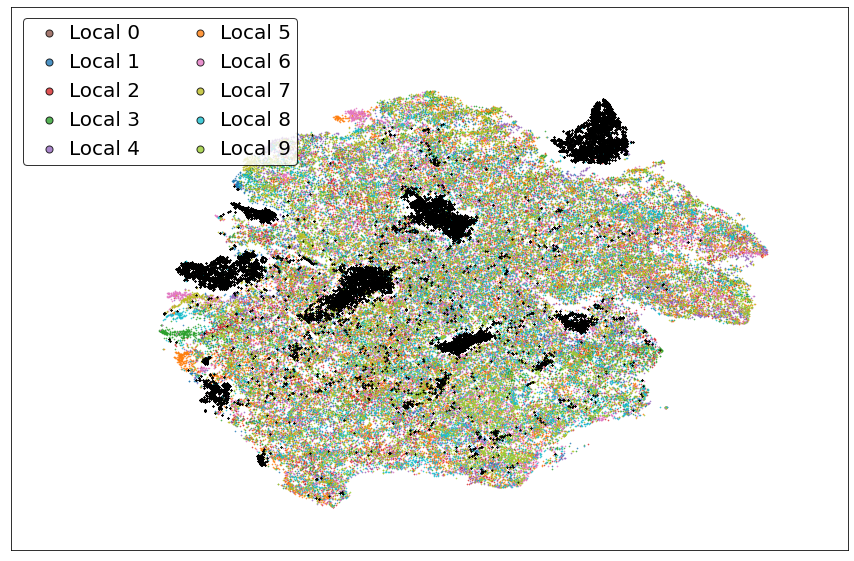}
        \caption[]%
        {{\small Homogeneous Locals (colored by \underline{\textit{classes}}})}
        \label{fig:tsne_homogeneous_local}
    \end{subfigure}
    \hfill
    \begin{subfigure}[b]{0.43\textwidth}  
        \centering 
        \includegraphics[width=\textwidth]{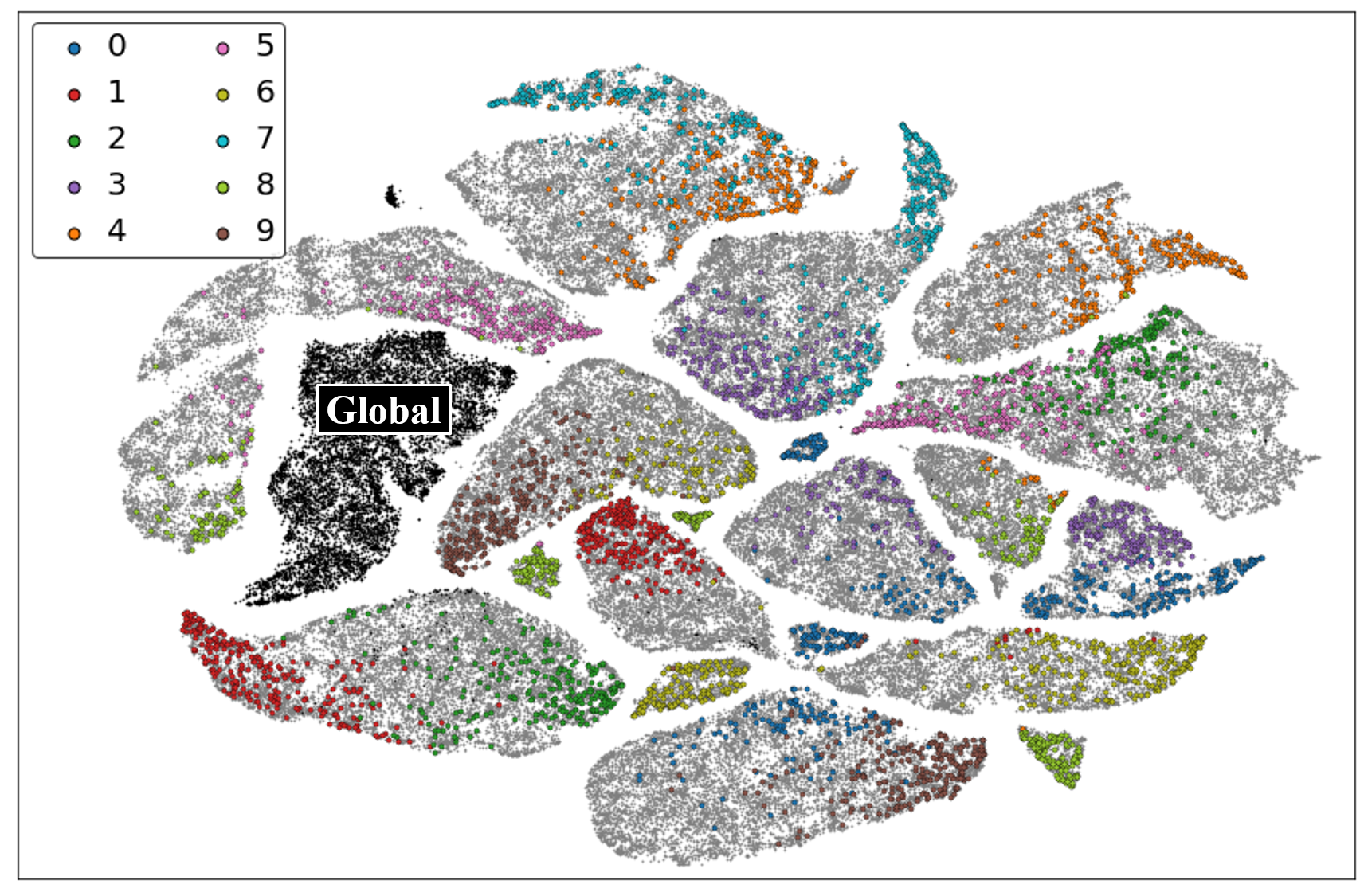}
        \caption[]%
        {{\small Heterogeneous Locals (colored by \underline{\textit{locals}}})}
        \label{fig:tsne_heterogeneous_local}
    \end{subfigure}
    \hfill
    \caption{\small T-SNE visualization of feature region shifting on CIFAR-10 test samples after local training on (a) Homogeneous local distributions and (b) heterogeneous local distributions. The T-SNE is conducted together for the test sample features of global and 10 local models.}
    \label{fig:local_vis2}
\end{figure*}
\vspace{-6pt}
\begin{figure*}[ht!]
    \centering
    \hfill
    \begin{subfigure}[b]{0.43\textwidth}  
        \centering 
        \includegraphics[width=\textwidth]{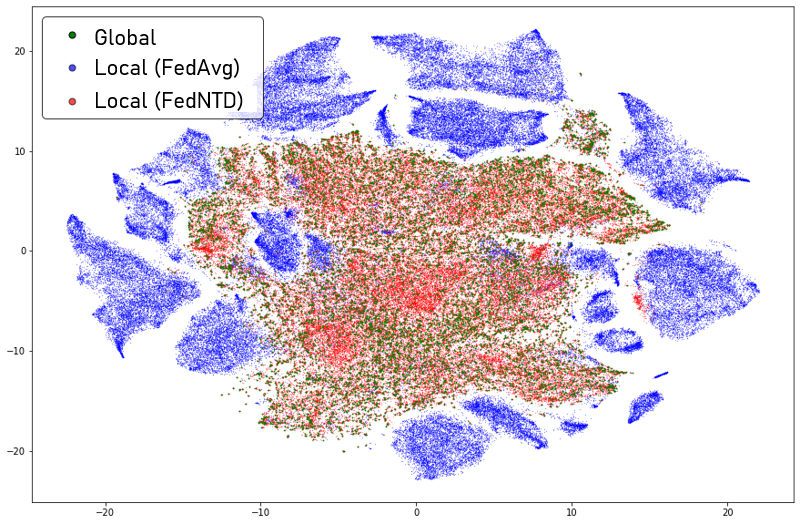}
        \caption[]%
        {{\small colored by different \textit{algorithms}}}
    \end{subfigure}
    \hfill
    \begin{subfigure}[b]{0.43\textwidth}  
        \centering 
        \includegraphics[width=\textwidth]{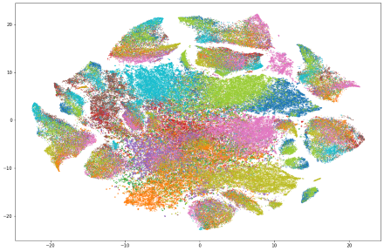}
        \caption[]%
        {{\small colored by \textit{classes}}}
    \end{subfigure}
    \caption[]
    {\small T-SNE on CIFAR-10 testset samples fter local training on \textit{heterogeneous} local distributions by \textbf{FedAvg} and \textbf{FedNTD}. The T-SNE is conducted together for the test sample features of global and 20 local models (10 for FedAvg and 10 for FedNTD).}
    \label{fig:tsne_avg_ntd}
\vspace{-12pt}
\end{figure*}

\clearpage
\section{Proof of Proposition 1}
\label{proof_proposition1}
\begin{proof}
Since the class-wise gradient $g_i$ are mutually orthogonal and have uniform weight, from the unitarily invariance of the 2-norm we have:
\begin{align}
\Lambda(\beta) &= \frac{\frac{1}{K} \sum_{k=1}^K  \Vert p^k + \beta\tilde{p}^k\Vert^2}{\Vert \sum_{k=1}^K p^k + \beta\tilde{p}^k\Vert^2 } \stackrel{(\spadesuit)}{=} \frac{1}{K} \frac{\sum_{k=1}^K \sum_{c=1}^\mathcal{C} (p^k(c) + \beta\tilde{p}^k(c))^2}{\sum_{c=1}^\mathcal{C} (\sum_{k=1}^K p^k(c) + \beta\tilde{p}^k(c))^2}\\
& \stackrel{(\clubsuit)}{=} \frac{1}{(1+\beta)^2}\frac{\mathcal{C}}{K^3}\sum_{k=1}^K \sum_{c=1}^\mathcal{C} (p^k(c) + \beta\tilde{p}^k(c))^2\\
& = \frac{1}{(1+\beta)^2}\frac{\mathcal{C}}{K^3}\sum_{k=1}^K \mathcal{C}\, \mathbb{E}_{c\in[\mathcal{C}]}[(p^k+\beta\tilde{p}^k)^2] = \frac{1}{(1+\beta)^2}\frac{\mathcal{C}^2}{K^3}\sum_{k=1}^K  \mathbb{E}_{c\in[\mathcal{C}]}[(p^k+\beta\tilde{p}^k)^2]\\
& = \frac{1}{(1+\beta)^2}\frac{\mathcal{C}^2}{K^3} \sum_{k=1}^K \left( \text{Var}_{c\in[\mathcal{C}]}[p^k+\beta\tilde{p}^k] + (1+\beta)^2 \right)\\
&= \frac{1}{(1+\beta)^2}\frac{\mathcal{C}^2}{K^3} \sum_{k=1}^K \left( \text{Var}_{c\in[\mathcal{C}]}\left[p^k+\beta\frac{1-p^k}{\mathcal{C}-1}\right]\right) +\frac{\mathcal{C}^2}{K^2}\\
& = \frac{1}{(1+\beta)^2}\frac{\mathcal{C}^2}{K^3} \sum_{k=1}^K \left(\big( 1 - \frac{\beta}{\mathcal{C}-1}\big)^2\text{Var}_{c\in[\mathcal{C}]}[p^k]\right) +\frac{\mathcal{C}^2}{K^2}\,.\label{prop1eq1}
\end{align}
Where the $(\spadesuit)$ follows from $p^k=\sum_{c=1}^\mathcal{C} p^k(c)\mathbf{e}_k(\in \Delta_\mathcal{C})$, and $(\clubsuit)$ holds because we are assuming uniform global data distribution. That is, we have the following equation from the symmetry over classes.
\begin{equation}
\sum_{k=1}^K p^k(c) = \sum_{k=1}^K \tilde{p}^k(c) = \frac{K}{\mathcal{C}}\,.
\end{equation}
By differentiating the equation~\eqref{prop1eq1}, we have:
\begin{align}
\frac{\partial \Lambda(\beta)}{\partial \beta} &= \left(\frac{\mathcal{C}^2}{K^3 (\mathcal{C}-1)^2} \sum_{k=1}^K \text{Var}_{c\in[\mathcal{C}]}[p^k]\right) \frac{\partial}{\partial \beta} \frac{(\mathcal{C}-(1+\beta))^2}{(1+\beta)^2}\\
&= \left(\frac{\mathcal{C}^2}{K^3 (\mathcal{C}-1)^2} \sum_{k=1}^K \text{Var}_{c\in[\mathcal{C}]}[p^k]\right) \left( -2 \frac{\mathcal{C}^2}{(1+\beta)^3} + 2\frac{\mathcal{C}}{(1+\beta)^2} \right)\\
&= -\left(\frac{2\mathcal{C}^3}{K^3 (\mathcal{C}-1)^2} \sum_{k=1}^K \text{Var}_{c\in[\mathcal{C}]}[p^k]\right) \left(\frac{\mathcal{C}}{(1+\beta)^3} - \frac{1}{(1+\beta)^2} \right)\,.
\end{align}
\end{proof}
\vspace{-18pt}
By defining $M_{K,\mathcal{C},p} >0$ in the first bracket, we have:
\begin{equation}
\frac{\partial \Lambda}{\partial \beta} = -M_{K,\mathcal{C},p} \left(\frac{\mathcal{C}}{(1+\beta)^3} - \frac{1}{(1+\beta)^2} \right)\,,
\end{equation}
for all $\beta\geq 0$. If $\beta \leq \mathcal{C}/2 - 1$, we have $\mathcal{C}/(1+\beta) \geq 2$, and get following desired inequality:
\begin{equation}
\frac{\partial \Lambda}{\partial \beta} \leq -M_{K,\mathcal{C},p}\frac{1}{(1+\beta)^2}\,.
\end{equation}

\section{Proof of Proposition 2}
\label{proof_proposition2}
\begin{proof}
First, we show the first equation. The summation for the true class is:
\begin{equation}
\mathcal{L}_{\textup{KL}}^{\textup{true}}=-\frac{1}{N}\sum_{i=1}^{N} q_{\tau}^{g,i}(y_i) \log\left[\frac{q_{\tau}^{l,i}(y_i)}{q_{\tau}^{g,i}(y_i)}\right]
\end{equation}
Note that $\sum_{i=1}^{N} = \sum_{c=1}^{\mathcal{C}}\sum_{i\in\mathcal{S}_c}$ and $i\in\mathcal{S}_c \Rightarrow y_i =c$. By using these, we get:
\begin{align}
\mathcal{L}_{\textup{KL}}^{\textup{true}}&=-\frac{1}{N}\sum_{i=1}^{N} q_{\tau}^{g,i}(y_i) \log\left[\frac{q_{\tau}^{l,i}(y_i)}{q_{\tau}^{g,i}(y_i)}\right] = -\frac{1}{N}\sum_{c=1}^{\mathcal{C}}\sum_{i\in\mathcal{S}_c}q_{\tau}^{g,i}(c) \log\left[\frac{q_{\tau}^{l,i}(c)}{q_{\tau}^{g,i}(c)}\right]\\
&=-\sum_{c=1}^\mathcal{C} \: \boldsymbol{p_c}  \cdot \left( \sum_{i\in\mathcal{S}_{c}} \frac{1}{|\mathcal{S}_{c}|}\: q_{\tau}^{g,i}(c) \log\left[\frac{q_{\tau}^{l,i}(c)}{q_{\tau}^{g,i}(c)}\right]\right)\\
&= -\sum_{c=1}^\mathcal{C} \: \boldsymbol{p_c}  \cdot \mathbb{E}_{i\in\mathcal{S}_{c}} \left[q_{\tau}^{g,i}(c) \log\left[\frac{q_{\tau}^{l,i}(c)}{q_{\tau}^{g,i}(c)}\right]\right] .
\end{align}
Next, we derive the not-true part of the Kullback-Leibler divergence:
\begin{equation}
-N{\mathcal{L}}_{\text{KL}}^{\text{not-true}}=\sum_{i=1}^{N}\sum_{c'\neq y_i}^{\mathcal{C}} q_{\tau}^{g,i}(c') \log\left[\frac{q_{\tau}^{l,i}(c')}{q_{\tau}^{g,i}(c')}\right]\,.
\end{equation}
By using the double summation technique ($\bigstar$), we have:
\begin{align}
-N{\mathcal{L}}_{\text{KL}}^{\text{not-true}}&=\sum_{c=1}^{\mathcal{C}}\sum_{i\in\mathcal{S}_c} \sum_{c'\neq c}^{\mathcal{C}} q_{\tau}^{g,i}(c') \log\left[\frac{q_{\tau}^{l,i}(c')}{q_{\tau}^{g,i}(c')}\right]=\sum_{c=1}^{\mathcal{C}}\sum_{c'\neq c}^{\mathcal{C}}\sum_{i\in\mathcal{S}_c} q_{\tau}^{g,i}(c') \log\left[\frac{q_{\tau}^{l,i}(c')}{q_{\tau}^{g,i}(c')}\right]\\
&\stackrel{(\bigstar)}{=}\sum_{c'=1}^{\mathcal{C}}\sum_{c\neq c'}^{\mathcal{C}}\sum_{i\in\mathcal{S}_c}  q_{\tau}^{g,i}(c') \log\left[\frac{q_{\tau}^{l,i}(c')}{q_{\tau}^{g,i}(c')}\right]=\sum_{c=1}^{\mathcal{C}}\sum_{c'\neq c}^{\mathcal{C}}\sum_{i\in\mathcal{S}_{c'}}  q_{\tau}^{g,i}(c) \log\left[\frac{q_{\tau}^{l,i}(c)}{q_{\tau}^{g,i}(c)}\right]\\
&=(\mathcal{C}-1)\sum_{c=1}^{\mathcal{C}}\frac{\sum_{c'\neq c} |\mathcal{S}_c'|}{\mathcal{C}-1} \left(\sum_{c'\neq c} \frac{1}{\sum_{c'\neq c}|\mathcal{S}_{c'}|} {\sum_{i\in\mathcal{S}_c'} q_{\tau}^{g,i}(c) \log\left[\frac{q_{\tau}^{l,i}(c)}{q_{\tau}^{g,i}(c)}\right]}\right)\\
&=(\mathcal{C}-1)\sum_{c=1}^{\mathcal{C}}N\boldsymbol{\tilde{p}_c} \: \cdot \mathbb{E}_{i\notin\mathcal{S}_{c}} \left[q_{\tau}^{g,i}(c) \log\left[\frac{q_{\tau}^{l,i}(c)}{q_{\tau}^{g,i}(c)}\right]\right]\,.
\end{align}
\end{proof}
Therefore, we get our desired result:
\begin{equation}
\frac{{\mathcal{L}}_{\text{KL}}^{\text{not-true}}}{\mathcal{C}-1} = \sum_{c=1}^{\mathcal{C}}\boldsymbol{\tilde{p}_c}\mathbb{E}_{i\notin\mathcal{S}_{c}} \left[q_{\tau}^{g,i}(c) \log\left[\frac{q_{\tau}^{l,i}(c)}{q_{\tau}^{g,i}(c)}\right]\right]\,.
\end{equation}

\section{Derivation of \autoref{eq:upper_bound}}
\label{proof_eq18}

\begin{proof}
The main part of the proof is well-known inequality for smooth functions, which is derived from the Taylor approximation. Since $\mathcal{L}_i:\mathcal{W}\subset\mathbb{R}^n\rightarrow\mathbb{R}$ is smooth function, we have
\begin{align}
\mathcal{L}_i(w)&=\mathcal{L}_i(w_i) + \nabla \mathcal{L}_i(w_i) \cdot (w-w_i) + \int_0^1 (1-t)(w-w_i)^\top \cdot \nabla^2 \mathcal{L}_i(w_i + t(w-w_i)) \cdot (w-w_i) dt\\
&=\mathcal{L}_i(w_i) + \int_0^1 (1-t)(w-w_i)^\top \cdot \nabla^2 \mathcal{L}_i(w_i + t(w-w_i)) \cdot (w-w_i) dt\\
&\leq \mathcal{L}_i(w_i) + \lambda \int_0^1 (1-t)(w-w_i)^\top \cdot (w-w_i) dt\tag{$\nabla^2\mathcal{L}_i(w) \preceq \lambda$}\\
&=\mathcal{L}_i(w_i) + \frac{\lambda }{2}\lVert w-w_i\rVert^2\,.
\end{align}
\end{proof}

\section{Proof of Proposition 3}
\label{proof:proposition3}
\begin{proof}
To show this corollary, enough to show that the below minimax problem is attained on the uniform distribution. 
\begin{equation}
\inf_{p\in \Delta_{\mathcal{C}}}\sup_{\mathbb{P}\in\Pi} \mathbb{E}_{p'\sim\mathbb{P}}[\lVert p' - p \rVert].
\end{equation}
Let us define $p \mapsto \sup_{\mathbb{P}\in\Pi} \mathbb{E}_{p'\sim\mathbb{P}}[\lVert p' - p \rVert]$ as $F(p)$. First, we check the continuity of $F$. That is:
\begin{align}
|F(p_2)-F(p_1)|\leq&\left| \sup_{\mathbb{P}\in\Pi} \mathbb{E}_{p'\sim\mathbb{P}}[\lVert p' - p_2 \rVert] - \sup_{\mathbb{P}\in\Pi} \mathbb{E}_{p'\sim\mathbb{P}}[\lVert p' - p_1 \rVert]\right|\\
\leq& \sup_{\mathbb{P}\in\Pi} \bigg| \mathbb{E}_{p'\sim\mathbb{P}}[\lVert p' - p_2 \rVert - \lVert p' - p_1 \rVert]\bigg| \leq \sup_{\mathbb{P}\in\Pi} \mathbb{E}_{p'\sim\mathbb{P}}[|\lVert p' - p_2 \rVert - \lVert p' - p_1 \rVert|]\\
\leq& \sup_{\mathbb{P}\in\Pi} \mathbb{E}_{p'\sim\mathbb{P}}[\lVert p_1- p_2 \rVert] \leq \lVert p_1- p_2 \rVert\,.
\end{align}
Therefore, since the function $F$ is 1-Lipschitz, it is clearly continuous. Now, since $\Delta_\mathcal{C}$ is compact, we have a minimizer $p_0\in \Delta_\mathcal{C}$ of above minimax value. Since norm and expectation is convex function, $F$ is convex. Therefore, for arbitrary minimizer $p_0$ and cycle $\sigma=(1\:2\:\cdots\:C)\in\mathcal{S}_{\mathcal{C}}$, we have:
\begin{equation}
\label{eqn:cor2-1}
F(\text{unif. dist}) = F\left(\frac{1}{\mathcal{C}} \sum_{i} \sigma^i(p_0)\right) \leq \frac{1}{\mathcal{C}} \sum_{i} F(\sigma^i(p_0))\,.
\end{equation}
Now, we argue that $F(\sigma^i(p_0)) = F(p_0)$. From the definition of $F$,
\begin{align}
F(\sigma^i (p_0)) &= \sup_{\mathbb{P}\in\Pi} \mathbb{E}_{p'\sim\mathbb{P}}[\lVert p' - \sigma^i(p_0) \rVert] = \sup_{\mathbb{P}\in\Pi} \mathbb{E}_{p'\sim\mathbb{P}}[\lVert \sigma^{i}(\sigma^{-i}(p')) - \sigma^i(p_0) \rVert]\\
&= \sup_{\mathbb{P}\in\Pi} \mathbb{E}_{p'\sim\mathbb{P}}[\lVert \sigma^{i}(\sigma^{-i}(p')) - \sigma^i(p_0) \rVert]\\
&= \sup_{\mathbb{P}\in\Pi}\int_{\Delta_\mathcal{C}}\lVert \sigma^{i}(\sigma^{-i}(p')) - p_0 \rVert\,d\mathbb{P}(p')\\
&= \sup_{\mathbb{P}\in\Pi}\int_{\Delta_\mathcal{C}}\lVert \sigma^{-i}( p') - p_0 \rVert\,d\mathbb{P}( \sigma^i(\sigma^{-i}(p')))\\
&= \sup_{\mathbb{P}\in\Pi}\int_{\Delta_\mathcal{C}}\lVert p'' - p_0 \rVert\,d\mathbb{P}(\sigma^i(p''))\\
&= \sup_{\mathbb{P}\in\Pi}\int_{\Delta_\mathcal{C}}\lVert p'' - p_0 \rVert\,d\mathbb{P}(p'') \tag{$\Pi$ is $\mathcal{S}_{\mathcal{C}}$-invariant ($\sigma^i \in \mathcal{S}_{\mathcal{C}}$)}=F(p_0)\,.\\
\end{align}
From the equation equation~\eqref{eqn:cor2-1}, we have:
\begin{equation}
F(\text{unif. dist}) \leq \frac{1}{\mathcal{C}} \sum_{i} F(\sigma^i(p_0)) = \frac{1}{\mathcal{C}} \sum_{i} F(p_0)  = F(p_0)\,.
\end{equation}
Since $p_0$ is minimizer, we can argue that the uniform distribution also attains the minimum.
\end{proof}
}

\end{CJK}
\end{document}